\documentclass[lettersize,journal]{IEEEtran}
\usepackage{algorithmic}
\usepackage{array}
\usepackage[caption=false,font=small,labelfont=sf,textfont=sf]{subfig}
\usepackage{textcomp}
\usepackage{stfloats}
\usepackage{url}
\usepackage{verbatim}
\usepackage{graphicx}

\usepackage{subfloat}
\usepackage{times}
\usepackage{amsmath,amssymb,amsopn,amstext,amsfonts}
\usepackage{cancel}
\usepackage[space]{cite}
\usepackage{pdfsync}
\usepackage{balance}
\usepackage{color}
\usepackage{mathtools}
\usepackage[ruled,vlined]{algorithm2e}
\usepackage{bm}

\usepackage{diagbox}
\usepackage{float}
\usepackage{epstopdf}
\usepackage{pifont}
\usepackage{fixltx2e}
\usepackage{amsmath}
\usepackage{multirow}
\usepackage{booktabs}
\usepackage{svg}
\usepackage{threeparttable}
\usepackage{caption}
\usepackage{lipsum}
\usepackage[linkcolor=black,citecolor=black,urlcolor=black,colorlinks=true]{hyperref}
\usepackage{makecell}
\usepackage{ulem}

\newcommand{\tp}{^{\mathrm{T}}}
\newcommand{\df}[1]{\mathrm{d}{#1}}
\newcommand{\rbrac}[1]{({#1})}
\newcommand{\norm}[1]{\Vert{#1}\Vert}

\def\BibTeX{{\rm B\kern-.05em{\sc i\kern-.025em b}\kern-.08em
    T\kern-.1667em\lower.7ex\hbox{E}\kern-.125emX}}

\begin{document}
\captionsetup{font={small}}
\title{Catch Planner: Catching High-Speed \\Targets in the Flight}
\author{
	{Huan Yu$^{*,1,2,3}$, Pengqin Wang$^{*,4}$, Jin Wang$^\dag$$^{,1,2,3}$, Jialin Ji$^{5,6}$, Zhi Zheng$^{1,2,3}$, Jie Tu$^{1,2,3}$, \\Guodong Lu$^{1,2,3}$, Jun Meng$^{3,7}$, Meixin Zhu$^{8}$, Shaojie Shen$^{4}$ and Fei Gao$^\dag$$^{,5,6}$\\}
	\vspace{-0.50cm}
\thanks{This work was supported in part by the National Natural Science Foundation of China under Grant 52175032, the ``Pioneer" and ``Leading Goose" R\&D Program of Zhejiang under Grant 2023C01070, and Robotics Institute of Zhejiang University under Grant K12107 and K11805.}
\thanks{
	$^{1}$The State Key Laboratory of Fluid Power and Mechatronic Systems,	School of Mechanical Engineering, Zhejiang University, Hangzhou 310027, China. $^{2}$the Engineering Research Center for Design Engineering and Digital Twin of Zhejiang Province, School of Mechanical Engineering, Zhejiang
	University, Hangzhou 310027, China. $^{3}$Robotics Institute of Zhejiang University, Hangzhou 310027, China. $^{4}$The Hong Kong University of Science and Technology (HKUST). $^{5}$ The State Key Laboratory of Industrial Control Technology, College of Control Science and Engineering, Zhejiang University, Hangzhou 310027, China. $^{6}$Huzhou Institute of Zhejiang University, Huzhou 313000, China. $^{7}$School of Electrical Engineering, Zhejiang University, Hangzhou 310027, China. $^{8}$The Hong Kong University of Science and Technology (Guangzhou).}
\thanks{Email: {\tt\small\{h.yu, dwjcom\}@zju.edu.cn}}
\thanks{\dag Corresponding author: Jin Wang, Fei Gao. }
\thanks{*These authors contributed to the work equally. }
}
\maketitle

\begin{abstract}
	Catching high-speed targets in the flight is a complex and typical highly dynamic task. However, existing methods require manual setting of catching height or time, resulting in lacks of adaptability and flexibility and cannot deal with multiple targets. To bridge this gap, we propose a planning-with-decision scheme called Catch Planner. For sequential decision making, a lightweight policy search method based on deep reinforcement learning is proposed. It is jointly trained with the motion planning and decoupled from physics to speed up training. For motion planning, we propose a trajectory optimization method that jointly optimizes the highly coupled catching time and terminal state. The core is the flexible-terminal constraint transcription. It converts the three unique constraints of catching into differentiable metrics, including equality constraints for terminal position and time, and inequality constraints that enable reasonable terminal position offset and attitude relaxation. In addition, sparse parameterization based on MINCO class considers both dynamic feasibility and collision avoidance constraints. As a result, a generally constrained quadrotor planning problem is transformed into an unconstrained optimization that can be solved reliably and efficiently. We also propose an online iterative optimization method for predicting differentiable trajectories of targets. Catch Planner provides a new paradigm for the combination of learning and planning, where all algorithms can be run in real time onboard at $100hz$. Extensive experiments are carried out in real-world and simulated scenes to verify the robustness and expansibility when facing a variety of high-speed flying targets.
\end{abstract}


\section{Introduction}
\IEEEPARstart{A}{utonomous} aerial robots, thanks to its high maneuverability, can be employed for many highly complex and dynamic tasks such as aerial interception \cite{cite:interception}, aerial perching \cite{cite:perching}, task dispatching \cite{cite:dispatching}, juggling \cite{cite:juggling}, racing \cite{cite:racing}, etc.. Catching is the most complex and typical problem among all the highly dynamic tasks, which requires not only accurate target motion prediction and precise interception but also adaptability and flexibility in the catching moment. In addition, it is necessary to make decisions of catching time and sequence when facing multiple targets. For catching task, motion planning and decision making (planning-with-decision) are the most important components since they determine all the expected quadrotor states. This article aims to provide a lightweight solution for catching, which can also be referenced for other highly dynamic tasks.

In existing catching methods, the catching time and terminal state still need to be determined manually or through a large number of trials. To the best of our knowledge, there is no catching methods that can maintain full autonomy and and flight flexibility, especially when facing multiple targets. We summarize the requirements of catching tasks for high-speed targets as follow (\textbf{FLAF}):
\begin{figure}[t]
	\begin{center}
		\includegraphics[width=1.0\columnwidth]{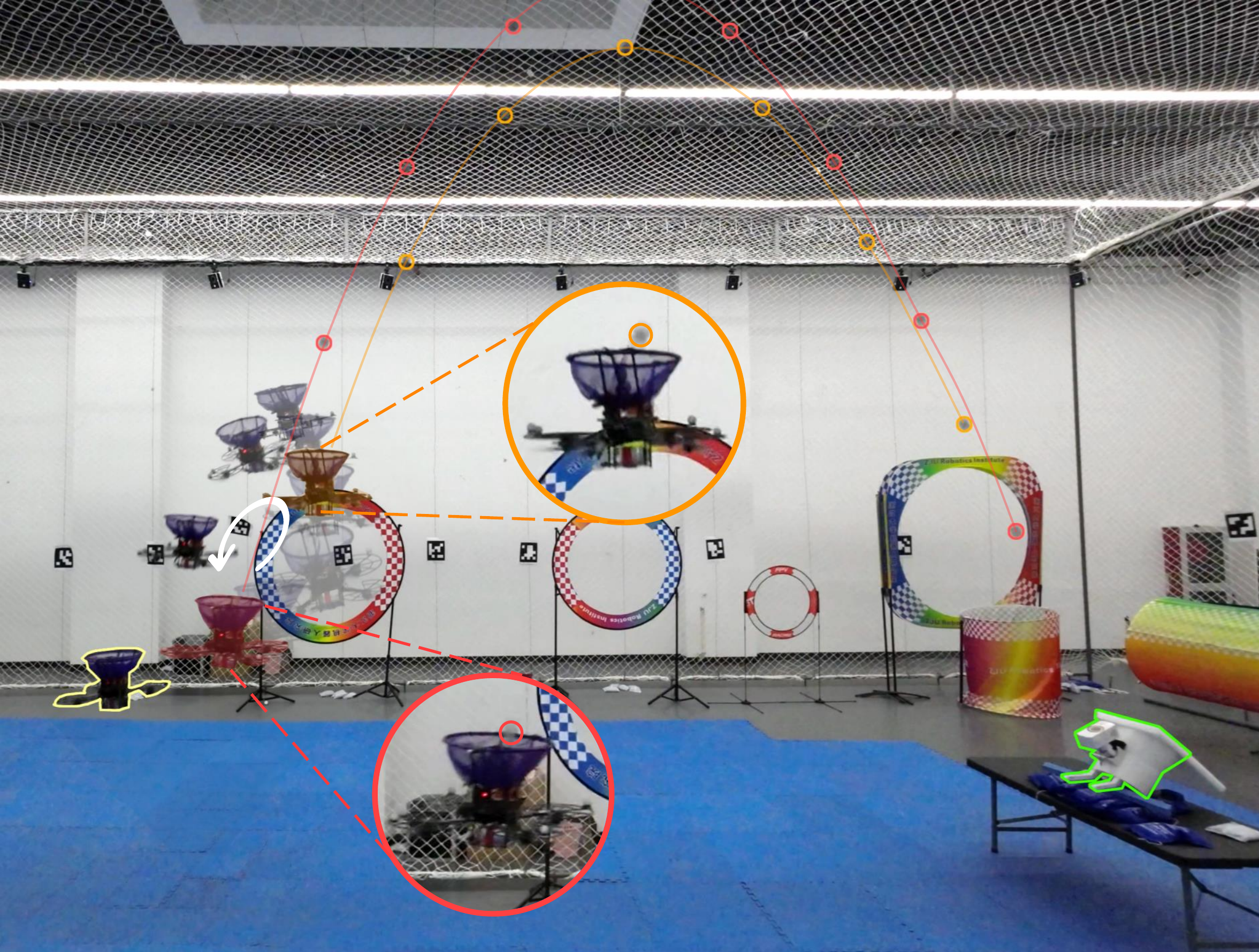}
	\end{center}
	\caption{
		\label{fig:introduction}
		The process of the quadrotor catching 2 flying targets, in which the throwing interval is 0.8$s$.  The targets trajectories are drawn in orange and red. The trajectory of the quadrotor is represented by several ghost images separated by 0.3$s$. The green box marks the pitching machine. The yellow box marks the starting position of quadrotor's motion planning.
	}
	\vspace{-0.5cm}
\end{figure}
\begin{itemize}
	\item \textbf{F}ormulistic: Planning-with-decision problems are hard or even impossible to explicitly formulate, which directly leads to the over simplification of planning-with-decision problems in previous work. Proper formulation is a powerful guarantee for optimal catching.
	\item \textbf{L}ightweight: Due to the errors in ego motion and target trajectory prediction, the drone needs high-frequency decision-making and re-planning, which requires lightweight method. 
	\item \textbf{A}daptive: When facing different flying targets, quadrotors need to adaptively determine highly coupled catching time and terminal position. 
	\item \textbf{F}lexible: Quadrotors should be allowed to catch targets in any reasonable attitude and deviations from the ideal catching position, rather than fixed.
\end{itemize}

Unfortunately, it is difficult, even internally contradictory, to achieve these four aspects at the same time. Planning is usually decoupled with decision making in previous methods. The over simplified decision making problem does not fully account for more refined trajectory planning. On the other hand, an excellent planning-with-decision method will iterate over all possible solutions and select the nearly optimal one rather than stop at a feasible solution. However, higher optimality comes from sophisticated formulation and more iterations or trials in the solution space, which significantly increases computation cost. Then, adaptability requires the joint optimization of coupled catching time and terminal position, which makes the problem non convex and challenging to find a solution. The introduction of flexibility further increases the difficulty of problem solving. In fact, only satisfying some basic requirements such as safety and feasibility while minimizing time and maximizing smoothness is already a difficult problem \cite{cite:ego_planner}. That is why most works are unable to take into account the above FLAF requirements at the same time.

\begin{figure}[t]
	\begin{center}
		\includegraphics[width=1.0\columnwidth]{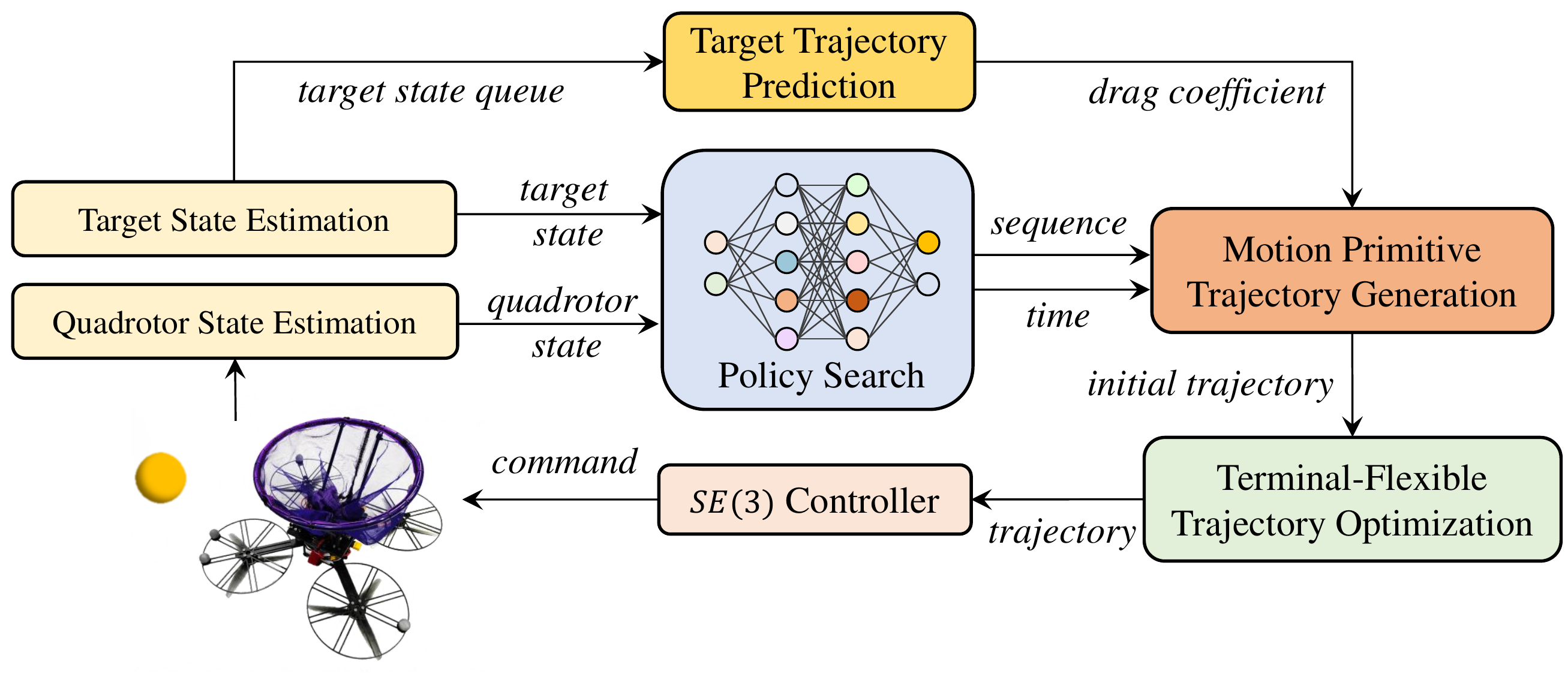}
	\end{center}
	\caption{
		\label{fig:frame}
		The overview of Catch Planner. All modules run online.
	}
	\vspace{-0.5cm}
\end{figure}

In this paper, we propose a systematic scheme called Catch Planner to meet FLAF demands and use table tennis ball as targets in our catching experiments. Catch Planner consists of Target Trajectory Prediction, Decision Making, Motion Primitive Generation and Trajectory Optimization modules, as shown in Fig. \ref{fig:frame}. We propose a deep reinforcement learning (DRL) based policy search method to solve the sequential decision making problem that is hard to formulate. In addition, we build simulation environment which decouples dynamics and physics to greatly reduce training computation consumption.(see Sec. \ref{section:decision-making}). Furthermore, the policy search results are used to generate control effort optimal motion primitive trajectory (see Sec. \ref{section:primitive}). Finally, we adopt MINCO trajectory class \cite{cite:MINCO} to conduct trajectory re-parameterization and optimization. We also propose a differentiable target trajectory prediction method with online iterative correction (see Sec. \ref{section:prediction}). Based on above, we jointly optimize highly coupled terminal states and catching time. Furthermore, we propose a lightweight terminal constraint transcription method enabling the quadrotor to catch targets at any reasonable attitude and position bias. Benefiting from sparse parametric optimization in MINCO and our constraint elimination and transcription approach (see Sec. \ref{section:optimization}), FLAF requirements are satisfied. Eventually, the robustness is verified by extensive experiments in real world and simulation. 

We summarize the contributions of our proposed Catch Planner as follows.
\begin{itemize}
	\item We propose a lightweight learning based sequential decision making method which is jointly trained with motion planning and decoupled with physics to accelerate training. The running time onboard is within $1ms$.
	\item We propose a terminal-flexible trajectory optimization method. The core is the constraint transcription to jointly optimize the coupled terminal state and catching time and eliminate inequality constraints that allow catching position offset and attitude relaxation.
	\item Catch Planner provides a new paradigm for merging learning and planning, benefiting from the accuracy of motion planning and the lightweight of neural network. The effectiveness and robustness are verified in extensive simulations and real-world experiments.
\end{itemize}

\section{Related Work}
There are few researches on the catching task, especially for planning-with-decision. Most existing methods treat target catching as a state-to-state motion planning problem. \cite{cite:bvp} proposes a closed-form solution to generate motion primitives for catching. This method is efficient, but the catching height must be determined manually or through trials. Furthermore, the dynamic feasibility of the trajectory is not considered while planning. \cite{cite:juggling_Wei} designs a controller to track the trajectory using the method from \cite{cite:bvp} with high following accuracy. \cite{cite:interception_JFR}  proposes a three-dimensional optimal terminal velocity control guidance for multicopter intercepting maneuvering drone with equal maneuverability level. This low-order planning leads to unsmooth trajectory. \cite{cite:juggling} shows two quadrocopters cooperatively juggling a ball back-and-forth. The trajectory is caculated under small-angle assumption, but it does not take actuator saturation into account. A target prediction method is also proposed in \cite{cite:juggling} by integrating forward the current position and velocity. All the above artificially fix the catching time and position, leading to reduced catching flexibility. In fact, coupled catching terminal and time make it difficult to jointly optimize the trajectory. Our work well overcomes this difficulty. We also provide a method to enable reasonable attitude and position offset under the premise of successful catching.

There are also some catching works that tend to obtain complex decision variables, such as time and sequence. Although the sequential decision making problem can be modeled by introducing integer variables, it will cause a second-level computational burden \cite{cite:MIP1}\cite{cite:MIP2}, which is intolerable for the catching task. The arrival of DRL methods on robotics tasks \cite{cite:policy_search20} \cite{cite:minimum_time22} \cite{cite:policy_search22} bring hope to solve the problem online. DRL has the power of improving the policy when the agent is constantly interacting with the environment by trial and error. However, most successful cases appear in simulation and games \cite{cite:nature19} rather than in reality. For catching task, \cite{cite:learning_catch} proposes an end-to-end method of visual reaction in the context of catching balls with a drone in visually rich synthetic environments. However, making decisions directly in the control space is difficult to migrate to the real world. \cite{cite:tennis} brings a learning method into reality and builds a library of “hitting motions” to determine the best hitting motion. But the success rate of interception is too low, which is because the decision does not consider the feasibility of quadrotor motion planning. To sum up, imperfect catch is caused by improper model. The decision making and motion planning are not considered coupled, resulting in the decision results can not be well implemented. Our method solve the above contradiction well, and provides a good paradigm for solving planning-with-decision problems.

Other works focus on the catching structure design \cite{cite:structure}, control \cite{cite:mpc_catch}, visual detection \cite{cite:detection1}  \cite{cite:detection2}, and state estimation \cite{cite:Shen}. All these have not solved the planning-with-decision problems, so will not be introduced.

\section{Problem Statement}
In this section, we summarize the formulation of the planning-with-decision problem for catching. The dynamic model of quadrotor and target are also introduced.

\subsection{Quadrotor Dynamic Model}
The quadrotor is modeled as a rigid body with six degrees of freedom: linear translation $p \in \mathbb R^3$ and rotation $R \in \mathrm{SE}(3)$. Translational motion depends on the gravitational acceleration $\bar g$ as well as the control input thrust $\tilde f$. Rotational motion takes the body rate $\omega \in \mathbb R^3$ as input. The model is \cite{cite:model_quadrotor}
\begin{equation}
	\begin{cases}
		\tau = \tilde f R \mathbf e_3 / m, \\
		\ddot{p} = \tau - \bar{g} \mathbf e_3, \\
		\dot{R} = R \hat \omega,
	\end{cases}
\end{equation}
where $\tau$ denotes the thrust, $\mathbf e_i$ is the $i$-th column of $\mathbf I_3$ and $\hat \cdot$ is the skew-symmetric matrix form of the vector cross product.
\begin{figure}[t]
	\begin{center}
		\includegraphics[width=1.0\columnwidth]{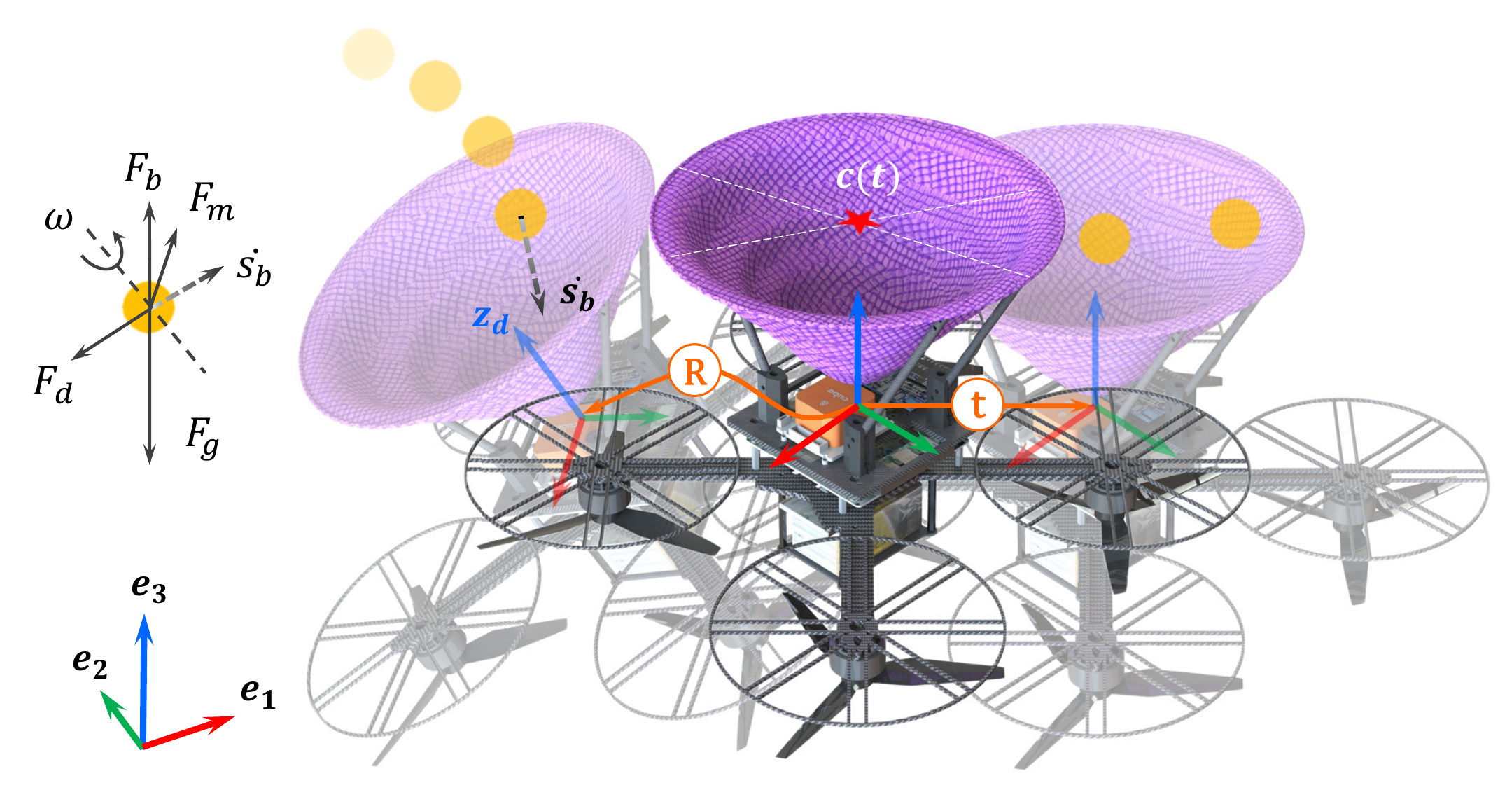}
	\end{center}
	\caption{
		\label{fig:model}
		Modeling overview and problem definition. The quadrotor can catch the targets in different attitudes (the target velocity $\dot{s_b}$ has an intersection angle with the catching net center $c(t)$), in which the movement is reflected by rotation $\mathbf{R}$. It can also catch the ball at different target positions (the catched position $s_b$ of the ball has a position offset with the catching net center $c(t)$) , in which the movement is reflected by translation $\mathbf{t}$.
	}
	\vspace{-0.5cm}
\end{figure}
Moreover, the catching loop with a net installed, shown in Fig. \ref{fig:model}, has a bias with the drone body. We use $\bar l$ to denote the length of the loop's centroid to the body. Thus the center of the loop is represented as $c(t) = p(t) - \bar l z_b(t)$, where $z_b(t)$ is the quadrotor attitude in time $t$.

\subsection{Target Dynamic Model}
The dynamic model of the flying target is high-order nonlinear which is affected by gravity $F_g$, aerodynamic drag $F_r$, Magnus force $F_m$, air buoyancy $F_b$, etc.. Since the ball rotation is small in this paper, the Magnus force is ignored. The air buoyancy is normally considered in conjunction with gravity. We simplify the ball's motion which includes $F_g$ and $F_r$ as \cite{cite:model_ball}
\begin{equation}\\
	\label{eq:ball_model}
	\ddot{s_b} = g - K_D \left\|  \dot{s_b} \right\|  \dot{s_b},
\end{equation}
where $\mathbf s_b$ denotes the ball's position. $K_D$ is proportionality coefficient. $\left\| \cdot  \right\| $ refers to the Euclidean norm. 

\subsection{Problem Formulation}
\label{subsection:problem_formulation}
The basic requirements of a planning trajectory $p(t)$ include dynamical feasibility and safety. Meanwhile, it is preferable to minimize control effort and time cost. The terminal state constraints are relaxed to catch more flexibly. It should also have decision making ability. In conclusion, the requirements of optimal catching give the following problem.
\begin{subequations}
	\label{eq:problem_formulation}
	\begin{align}
		\min_{p(t), T} & \label{eq:cost_function}~\sum_{i} \mathcal J_i = \sum_{i}\int_{0}^{T} {\norm{p_{i}^{(s)}(t)}^2} \df{t} + \rho T_i,                                                      \\
		s.t.~          & \label{eq:T_constrain} ~T_i > 0, \\
		& \label{eq:initial_final_constrain}~p_{0}^{(s-1)}(0)= \mathbf {p_0},     \\
		& \label{eq:continue_constrain}~p_{i}^{(s-1)}(0)=p_{i-1}^{(s-1)}(T_i),               \\
		& \label{eq:actuator_constrain}~\norm{D_{i}(t)}\leq D_{max}, ~\forall t\in[0,T],                                                                \\
		& \label{eq:ground_constrain}~\mathbf e_3^T p_{i}(t) \geq z_{min}, ~\forall t\in[0,T],                                                                           \\
		& \label{eq:catching_position}~p_{i}(T_{i})=s_b - \bar l \mathbf  z_d ~\text{(in the ideal case)},                                                 \\
		& \label{eq:catching_angle}~\left<-\dot{s_b}, \mathbf  z_d\right>\leq \theta_{max},              
	\end{align}
\end{subequations}
where  Eq. \ref{eq:cost_function} is a general form that trades off smoothness and aggression, with the goal of minimizing control effort and time cost. $T_i$  is the allocated time for each trajectory which is positive. Eq. \ref{eq:initial_final_constrain} denotes the initial state which is a zero matrix. Eq. \ref{eq:continue_constrain} provides a guarantee for the continuity of front and rear trajectories. Eq. \ref{eq:actuator_constrain} shows actuator constraints where $D$ denotes speed $p_{i}^{(1)}(t)$, body rate $\omega$ and thrust $\tau$ limitations. Eq. \ref{eq:ground_constrain} is a safety constraint to avoid hitting the ground, meaning that the quadrotor's position cannot be lower than a safe altitude. Eq. \ref{eq:catching_position} denotes the ideal catching position, which can be relaxed using our proposed method, meaning that the targets can fall anywhere within the net envelope, shown as Fig. \ref{fig:model}. $\mathbf  z_d = \tau / \norm{\tau}$ denotes the quadrotor’s desired catching attitude. It is restricted by Eq. \ref{eq:catching_angle}, meaning that the angle between $\mathbf  z_d$ and $\dot{s_b}$ needs to be within a safe angle to avoid the target hitting outside of the catching net.

Multiple trajectories are optimized together when catching multiple targets, which is represented as $\sum\limits _{i} \cdot$, also including the catching sequence. It is guaranteed by the learning based decision making. The goal is to maximize the cumulative reward. The optimal policy $\pi^*$ can be written explicitly for a discount factor $\gamma$:
\vspace{-0.2cm}
\begin{align}
	\pi^* = \mathop{\arg\max}\limits_{\pi} \mathbb{E}[\sum_{t=1}^T\gamma^{t-1}r_t].
\end{align}

The optimal policy determines the order of catching, which is implicitly expressed as $\sum\limits _{i} \cdot$ in Eq. \ref{eq:problem_formulation}.

\section{Learning-based Sequential Decision Making}
\label{section:decision-making}
In this section, we introduce the decision making framework and explain core terms design including rewards design, observation representation, action representation and invalid action masking in details. At the end we present the training strategy and environment.
\subsection{Learning Neural Network Policies}
\subsubsection{Decision Making Framework}
In our task, the decision making policy receives current quadrotor position, quadrotor attitude, quadrotor velocity, targets position and targets velocity as inputs, and outputs predicted time variables for catching each target. The decision making policy is trained using Proximal Policy Optimization (PPO) \cite{cite:ppo}, which demonstrates impressive performance for continuous tasks. Our main objective is to update policy search network parameters $\theta$ by maximizing the following:
\begin{align}
	\begin{split}
			\mathcal J_{DM}(\theta) =  \mathbb{E}_t[ &\min(\frac{\pi_\theta(a_t|s_t)}{\pi_{\theta_{old}}(a_t|s_t)}\hat{A_t}, \\
		&{\rm clip}(\frac{\pi_\theta(a_t|s_t)}{\pi_{\theta_{old}}(a_t|s_t)}, 1-\epsilon, 1+\epsilon)\hat{A_t})],
	\end{split}
\end{align}
where $\theta_{old}$ and $\theta$ are the policy parameters before and after the update respectively, $\hat{A_t}$ is an estimator of adavantage function at time step $t$ and $\epsilon$ is a hyperparameter.

\subsubsection{State Representation}
The following components which are essential and minimalistic to represent states are used:
\begin{align}
	\textbf s_t= \left [ p_d(t), q_d(t), p^{ (1) }_d(t), p_t(t), p^{(1)}_t(t) \right ],
\end{align}
where $p_d(t), q_d(t), p^{(1) }_d(t), p_t(t), p^{(1)}_t(t)$ stand for the quadrotor’s position, attitude represented by quaternion, velocity and targets’ position and velocity at time step $t$, respectively.

\subsubsection{Action Representation}
In our case, the decision making time and sequence results are used in trajectory generation. To achieve this goal, we define the action $\textbf a_t$ to be predicted cathing time sequence, which can be written as:
\begin{subequations}
\begin{align}
	\pi_\theta(\textbf s_t) &= \textbf a_t = \mathcal T_p, \\
	\pi_\theta(\textbf s_t) &\sim \mathcal N (\mu_t, \sigma_t),
\end{align}
\end{subequations}
where $\theta$ means the network parameters for policy search, $\mathcal T_p$ represents predicted catching time sequence and each element of the sequence shows the predicted catching time for each target, $\mathcal N$ is a multivariate Gaussian density function, $\mu_t$ and $\sigma_t$ are the mean and variance of the Gaussian distribution respctively.

\subsubsection{Reward Design}
In order to make the quadrotor catch as more targets as possible while flying smoothly and agilely, we design the reward signal which takes both caught targets amount, trajectory cost, optimized time and sequence into consideration at the same time. The reward signal $\textbf r_t$ is calculated by:
\begin{align}
	\textbf r_t = \lambda_n(e^{n(t)-N}-e^{-N})-\lambda_cln(\mathcal J)-\lambda_t||\mathcal T_p-\mathcal T_o||_2,
\end{align}
where $n(t)$ suggests caught targets amount at time step $t$, $N$ represents total targets amount, $\mathcal J$ indicates trajectory cost at time step $t$, $\lambda_n$ $\lambda_c$ and $\lambda_t$ are catching reward coefficient, trajectory cost punishment coefficient and time punishment coefficient respectively and $||\mathcal T_p-\mathcal T_o||_2$ means the L2 norm between predicted catching time sequence and optimized catching time sequence.

\subsubsection{Invalid Action Masking}
An action sampled directly from the whole action space can be typically invalid because time variable ranges from 0 to infinity. Invalid action masking is an applicable solution to deal with such problems. In our task, we firstly limit each time variable to the range from 0 to $t_{max}$, where $t_{max}$ demonstrates maximum flying time of targets. Furthermore, it is also hard for quadrotor to catch targets within a surprisingly short time such as 0.1s. Therefore, we limit the predicted catching time within the range from $t_{min}$ to $t_{max}$, where $t_{min}$ denotes the minimum expected time for the quadrotor to catch targets.

\vspace{-0.2cm}
\subsection{Training Strategy and Environment}
One hundred independent parallel training environments and agents are created and utilized to improve data collection speed, increase data quality and accelerate training speed. Our policy network is MLP with 2 hidden layers of 64 units and ReLU activation function, which is lightweight to run onboard within $1ms$ and used for trajectory generation and optimization at the back end.

To speed up the self-supervised neural network policies training, we decouple physics and motion planning. During training, the initial quadrotor position is set to be a random point in a $1m \times 1m \times 1m$ cube. The targets position and velocity are randomly initialized within a certain range and depend on simulation scenarios. The random state initialization enhances the generalization ability of the model, which ensures that the proposed method can deal with the targets of different speeds and positions. We define the process including targets states random initialization, quadrotor states random initialization, targets throwing, decision making, motion planning and targets catching as one step. After decision making, trajectory generation and optimization, we can directly obtain the catching result and trajectory cost by calculating the trajectory terminal state and target motion instead of executing the whole trajectory, which offers great impovement on training speed.

\begin{figure}[t]
	\begin{center}
		\includegraphics[width=0.8\columnwidth]{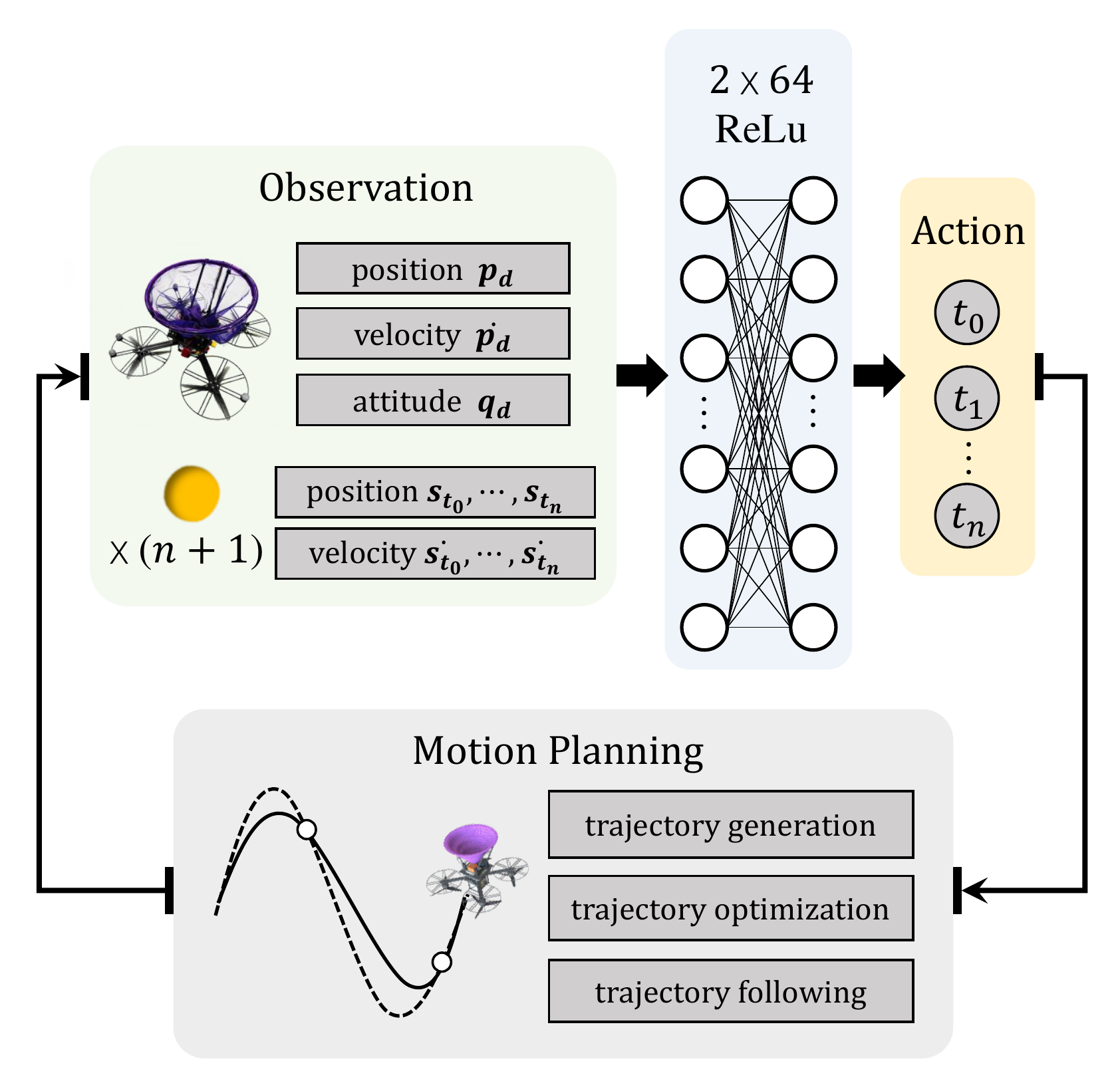}
	\end{center}
	\caption{
		\label{fig:learning}
		The overview of the learning based sequential decision making which costs $1ms$ onboard.
	}
\end{figure}

\section{Motion Primitive Trajectory Generation}
In this section, the purpose is to find a trajectory that is close to the optimal trajectory, considering requirements for catching. First, we introduce the analytic expression method of target trajectory prediction, in which the expression are optimized by online iteration. Then we introduce the method of control effort cost minimum trajectory generation, which uses the results of decision making module as input.

\subsection{Target Trajectory Prediction}
\label{section:prediction}
Catching task needs accurate target state estimation and trajectory prediction. Eq. \ref{eq:ball_model} shows that $\dot{s_b}$ brings nonlinear effects for ball flying dynamics model, which is unacceptable for trajectory optimization. \cite{cite:ball_simple_model} proposed a simplified linear target model and trained a parameter identification neural network to obtain $K_D$ offline. We extend this model to online. Ball’s linear equation of motion $s_b(t) $ can be expressed as 
	\begin{align}
		\label{eq:ball_simple_model}
		s(t) = &s_0 -  \dfrac{ \dot{s_0}+ \dfrac{g} { K_D}}{K_D} e^{-K_Dt-1}-\dfrac{g}{K_D}t,
	\end{align}
where $g$ and $K_D$ are decoupled in Euclidean space, $s_0$ is the the initial ball's position each update. Note that Eq. \ref{eq:ball_simple_model} is continuously differentiable, so it is easy to obtain the gradient $\partial s_b/\partial c_i$ and $\partial s_b/\partial T_i$. Then the extended  kalman filter \cite{cite:control_book} (EKF) is used to optimize the observation value. The state transformation function is presented as
\begin{align}
	\begin{bmatrix}
		\mathbf s_{b_i}^{t} \\
		\mathbf v_{b_i}^{t} 
	\end{bmatrix}
	=
	\begin{bmatrix}
		\mathbf I_{3\times 3} & \mathbf I_{3\times 3}\Delta t \\
		\mathbf0 & \mathbf I_{3\times 3}e^{-K_i \Delta t} \\
	\end{bmatrix} 
	\begin{bmatrix}
		\mathbf s_{b_i}^{t-1} \\
		\mathbf v_{b_i}^{t-1} 
	\end{bmatrix} 
	+
	\begin{bmatrix}
		\mathbf0 \\
		1
	\end{bmatrix}
	\frac{g(e^{-K_i \Delta t} - 1)}{K_i},
\end{align}
where $i \in \{ x, y, z\}$, $\begin{bmatrix} \mathbf s_{b_i}^{t}&\mathbf v_{b_i}^{t} \end{bmatrix}$ represent the target state at the moment $t$. Kalman state estimate at $t$ can be expressed by
\begin{align}
	\mathbf s_{b}^{cor} = \mathbf s_{b}^t + K_a(Z-\mathbf Hs_{b}^t),
\end{align}
where the observation matrix  $\mathbf H = \left [\mathbf I_{3\times 3}\   \mathbf 0_{3\times 3} \right ]^T$, $K_a$ denotes kalman gain. $ Z$ denotes the motion capture observation. To make $K_D$ more accurate, we use nonlinear least squares method \cite{cite:LSM} to optimize $K_D$ online. The cost function is constructed as
\begin{align}
	\mathcal J_{K_D} = \sum_{i}^{i+n}(\mathbf s_{b}^{cor}-\mathbf s_{b})^2,
\end{align}
where $n$ represents the number of states used for optimization in continuous time. The goal is to minimize it, which is efficiently solved by Ceres Solver \footnote{https://github.com/ceres-solver/ceres-solver}.

\subsection{Motion Primitive trajectory generation}
\label{section:primitive}
The motion primitive generator \cite{cite:bvp} is considered as an efficient method to plan a state-to-state trajectory, which is closed-form. We use it as the front end of trajectory optimization. Consider an $m$-dimensional trajectory whose $i$-th piece is denoted by a $N = 2s - 1$ degree polynomial:
\begin{align}
	\label{eq:trajectory}
	p(t)= \textbf{c}_i^T \beta (t), t \in \left[ 0, T_i \right],
\end{align}
where $\textbf{c}_i \in \mathbb{R}^{2s \times m}$ is the coefficient matrix, $T_i$ is obtained from the policy search, $\beta (t) = (1,t,\cdots, t^N)^T$ is the natural basis.

 Following \cite{cite:bvp}, we consider the motion of the quadrotor in terms of the jerk $s = 3$, allowing the system to be considered as a triple integrator in each axis, and minimize the cost function
\begin{align}
	\mathcal J_{MP} = \frac{1}{T} \int_{T}^{0}\left\| p^{(3)}(t) \right\|^2 dt,
\end{align}
where $T$ is from decision making module. Just like \cite{cite:bvp}, we use Pontryagin’s minimum principle \cite{cite:control_book} to generate the optimal state trajectory by introducing the costate $\lambda_m = (\lambda_1, \lambda_2, \lambda_3)$ and defining the Hamiltonian function $H(s, j, \lambda_m)$ as 
\begin{subequations}
	\begin{align}
		\label{eq:hamiltonian}~H(s, j, \lambda_m) &=\frac{1}{T}j^2+\lambda_m^Tf_s(s, j) = \frac{1}{T}j^2+\lambda_1v+\lambda_2a+\lambda_3j, \\
		\label{eq:bvp}\dot{\lambda_m} &=\triangledown H(s^\star, j^\star, \lambda_m) = (0, -\lambda_1, -\lambda_2),
	\end{align}
\end{subequations}
where $ j^\star$ and $ s^\star$ represent the optimal input and state. The cost value can be calculated as
\begin{align}
	\mathcal J_{MP} = \gamma^2 + \beta \gamma T + \frac{1}{3} \beta^2 T^2 + \frac{1}{3}\alpha \gamma T^2 + \frac{1}{4} \alpha \beta T^3 + \frac{1}{20} \alpha^2 T^4,
\end{align}
where $\alpha, \beta$  and $\gamma$ are constants. They are explicitly calculated according to current state of the quadrotor $s(0)$ and final state $s(T)$, which is determined by Eq. \ref{eq:ball_simple_model} and $T$ from the policy search. For more details, please see \cite{cite:bvp}. Then we can get the initial trajectory $p_0(t)$.

\section{Terminal-flexible Trajecory Optimization \\ for Cathing}
\label{section:optimization}
In this section, we summarize the proposed trajectory optimization method for catching, which jointly optimizes all the requirements in the planning module. The problem form can be efficiently transformed from Eq. \ref{eq:problem_formulation} into a new form of unconstrained nonlinear programming. The generated trajectory  $p_0(t)$ is used to generate the initial values. Moreover, a lightweight terminal constraint transcription method is proposed to make catching more flexible. 
\subsection{MINCO Trajectory Class}
We adopt $\mathfrak{T}_{\mathrm{MINCO}}$ for trajectory representation, which is a \textbf{min}imum \textbf{co}ntrol (MINCO) \cite{cite:MINCO} polynomial trajectory class, defined as
\begin{equation}
	\begin{aligned}
		\mathfrak{T}_{\mathrm{MINCO}} = \Big\{ & p(t):[0, T]\mapsto\mathbb{R}^m~ \Big|~\mathbf{c}=\mathcal{M}(\mathbf{q},\mathbf{T}),~ \\
		& ~~\mathbf{q}\in\mathbb{R}^{m(M-1)},~\mathbf{T}\in\mathbb{R}_{>0}^M\Big\},
	\end{aligned}
\end{equation}

where an $m$-dimensional trajectory $p(t)$ is represented by a piece-wise polynomial of $M$ pieces and $N=2s-1$ degree.  In this paper, we use $s=4$ for minimum snap for enough freedom of trajectory. All trajectories in $\mathfrak{T}_{\mathrm{MINCO}}$ have compact parameterization by only the intermediate waypoint vector $\mathbf{q}$ and time vector $\mathbf{T}$ via the linear-complexity formulation $\mathbf c = \mathcal M(\mathbf q, \mathbf T)$. Furthermore, any cost function can be calculated by $\mathcal J(\mathbf q, \mathbf T) = F(\mathcal{M}(\mathbf q, \mathbf T), \mathbf T)$. The mapping also gives a linear-complexity way to cpmpute $\partial \mathcal J / {\partial \mathbf q}$ and $\partial \mathcal J / {\partial \mathbf T}$ from $\partial F / {\partial \mathbf q}$ and $\partial F / {\partial \mathbf T}$. After that, a high-level optimizer is able to optimize the objective efficiently.

\subsection{Trajectory Joint Optimization}
\label{subsection:nonlinear_optimization}
Considering all described requirements in Eq. \ref{eq:problem_formulation}, we adopt the compact parameterization of $\mathfrak{T}_{\mathrm{MINCO}}$, temporal constraint elimination, and constraint penalty to transform trajectories generation problem into an unconstrained nonlinear optimization problem:
\begin{align}
	\label{eq:nonlinear_const_function}
	\underset{\mathbf c,\mathbf T}{min} \sum_{i+1} \mathcal J_e + \lambda \cdot \left [ \mathcal J_t, \mathcal J_p, \mathcal J_\theta  ,  \mathcal J_v, \mathcal J_\omega, \mathcal J_f, \mathcal J_g \right],
\end{align}
where $i$ is the number of the targets,  $\lambda$ is the weight vector. 

\subsubsection{Control Effort $\mathcal J_e$} The control cost  is the same as Eq. \ref{eq:cost_function}. Then the gradients $\partial{\mathcal J_e}/\partial c$ and $\partial{\mathcal J_e}/\partial T$ are evaluated as
\begin{subequations}
	\begin{align}
		\mathcal J_e = & \int_{0}^{T} {\norm{p_{i}^{(4)}(t)}^2} \df{t} , \\
		\frac{\partial{\mathcal J_e}}{\partial c_i} = & 2\left( \int_0^{T_i} \beta^{(3)}(t)\beta^{(3)}(t)\tp \df{t} \right) c_i, \\
		\frac{\partial{\mathcal J_e}}{\partial T_i} = & c_i\tp\beta^{(3)}(T_i)\beta^{(3)}(T_i)\tp c_i,
	\end{align}
\end{subequations}

\subsubsection{Temporal Constraint Elimination $\mathcal J_t$} In order to catch more targets, shorter total time is eagerly expected. We minimize the total time just like Eq. \ref{eq:cost_function} as 
\begin{align}
	\mathcal J_t = & \rho T_i.
\end{align}
Meanwhile, Eq. \ref{eq:T_constrain}, which requires the strict positiveness of each entry in $\mathbf T$, does harm to the optimization process. We eliminate it by explicit diffeomorphism in Euclidean spaces as $T_i = e^{t_i}$. Therefore,  $t_i \in \mathbb{R}$ instead of $T_i \in \mathbb{R_+}$ becomes the new optimal variable. The gradients are calculated by 
\begin{align}
	\frac{\partial \mathcal J_t}{\partial c_i} = 0, \frac{\partial \mathcal J_t}{\partial t_i} = \frac{\partial \mathcal J_t} {\partial T_i} e^{t_i}.
\end{align}

\subsubsection{Flexible Terminal Transformation $\mathcal J_p, \mathcal J_\theta$}
Since the flat-output \cite{cite:minsnap} characteristics of quadrotor, the terminal state can be determined by $p_i^{(s-1)}(T_i )$. Eq. \ref{eq:catching_position} shows the ideal situation for catching. Such a strict restriction is obviously not in line with reality. The quadrotor should be able to catch the ball at any position in the net with any attitude, as shown in the Fig. \ref{fig:model}. We relax the constraints by designing a penalty function
\begin{align}
	\label{eq:position_penalty}
	 \mathcal J_p = \mathcal L_\varepsilon  \left[ \mathcal G_p(t) \right], \mathcal G_p(t) = \norm{p-s_b- \bar l\mathbf z_d}.
\end{align}

\begin{figure}[t]
	\begin{center}
		\includegraphics[width=0.8\columnwidth]{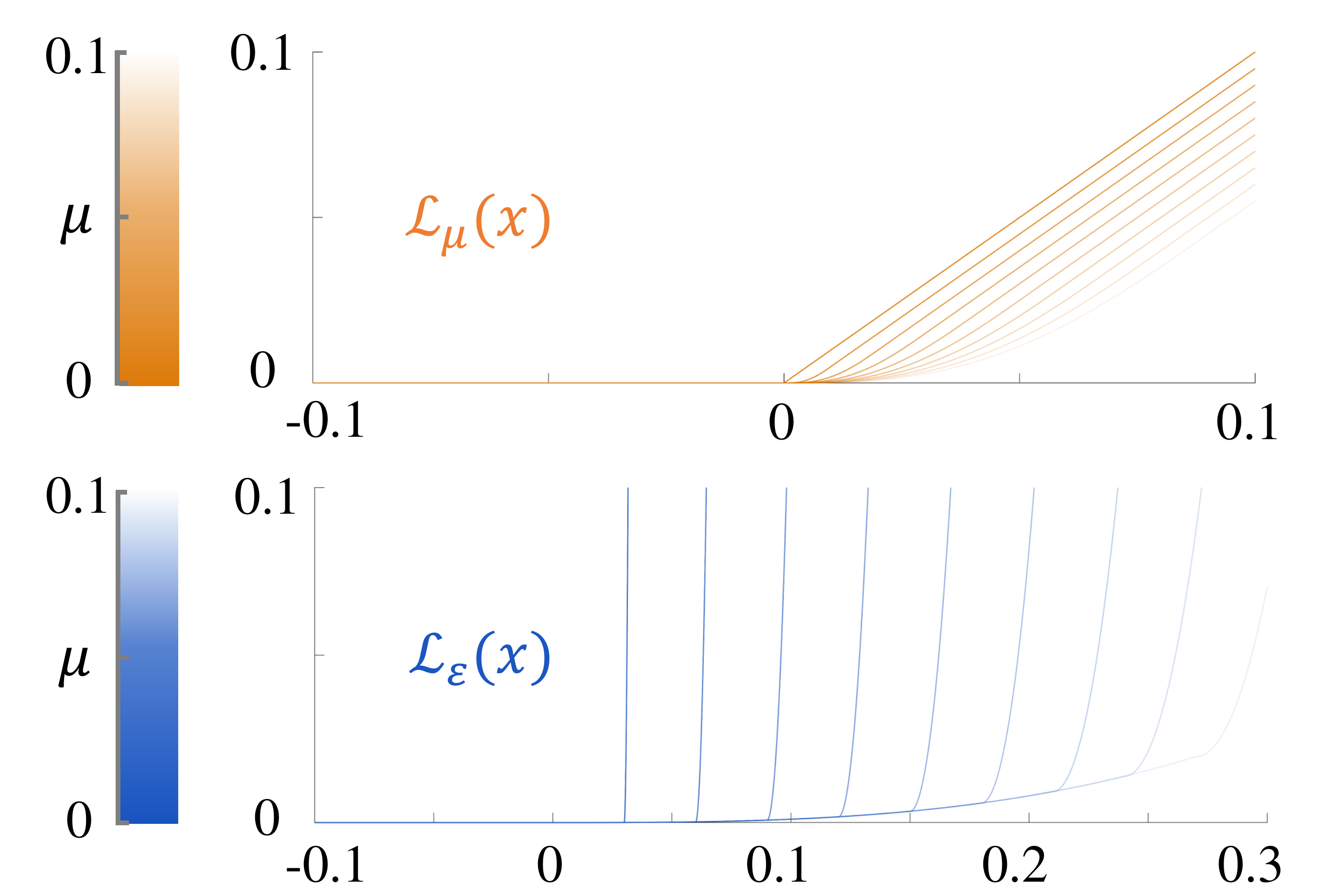}
	\end{center}
	\caption{
		\label{fig:logic}
		The smoothed function.
	}
	\vspace{-0.5cm}
\end{figure}

In order to make the catching position flexible while inside the net, inspired by \cite{cite:smoothed_function}, we design a differentiable and continuous smoothed function as
\begin{align}
	\label{eq:logistic_function}
	\mathcal L_\varepsilon [x] = 
	\begin{cases}
		0,                    & x \leq 0,       \\
		x^3, & 0 < x \leq \varepsilon, \\
		\varepsilon^3+(x-\varepsilon)^2x/\varepsilon^4,            & x > \varepsilon,
	\end{cases}
\end{align}
where the gradient changes smoothly when $0 < x \leq \varepsilon$, while drastic when $x > \varepsilon$, meaning stronger penalties for violating constraints. Then the gradients are obtained by
\begin{equation}
	\label{eq:feasible_p_gradient}
	\begin{aligned}
		\frac{\partial{\mathcal J_p}}{\partial c_i}= & \frac{\partial{\mathcal J_p}}{\partial{\mathcal G_p}} \frac{\partial{\mathcal G_p}}{\partial c_i} = 0, 	\frac{\partial{\mathcal J_p}}{\partial T_i}= &\frac{\partial{\mathcal J_p}}{\partial{\mathcal G_p}} \frac{\partial{\mathcal G_p}}{\partial s_b} \frac{\partial s_b}{\partial T_i}.
	\end{aligned}
\end{equation}
It is worth noting that each item can be calculated by Eq. \ref{eq:ball_simple_model}, Eq. \ref{eq:position_penalty} and Eq. \ref{eq:logistic_function}.

The angle constraint Eq. \ref{eq:catching_angle} requires that the quadrotor's terminal attitude $\mathbf z_d$ and the moving direction of the flying target are kept within a safe range. This constraint is translated into the following penalty function.
\begin{align}
	\label{eq:angle_penalty}
	\mathcal G_\theta(t) = \cos \theta_{safe}- \frac{\dot{s_b}\cdot \mathbf z_d}{\norm{\dot{s_b}}\norm{\mathbf z_d}},  \mathcal J_\theta = \mathcal L_\varepsilon \left[ \mathcal G_\theta(t) \right].
\end{align}
And its gradients can be caculated just like Eq. \ref{eq:feasible_p_gradient}.

\subsubsection{Continuous-Time Constraints $\mathcal J_v, \mathcal J_\omega, \mathcal J_f, \mathcal J_g$}
$\mathfrak{T}_{\mathrm{MINCO}}$ can be freely deformed to meet the continuous-time constraints $\mathcal G$. However, enforcing $\mathcal G$ over the entire trajectory involves infinitely many inequalities that cannot be solved by constrained optimization. Inspired by \cite{cite:MINCO}, We transform $\mathcal G$ into finite inequality constraints using integral of constraint violations.
\begin{subequations}
	\vspace{-0.4cm}
	\begin{align}
		\label{eq:time_integral_penalty}
		\mathcal I^\star_i=                                             & \frac{T_i}{\kappa}\sum_{j=0}^{\kappa_i}\bar{\omega}_j\mathcal{G}_\star \rbrac{\frac{j}{\kappa} T_i },      
		\mathcal J_\star = \sum_{i=1}^{M} \mathcal I^\star_i,                                                                                                                                                    \\
		\frac{\partial \mathcal J_\star}{\partial c_i} =              & \frac{\partial \mathcal I^\star_i}{\partial \mathcal G_\star} \frac{\partial \mathcal G_\star}{\partial c_i},
		~\frac{\partial \mathcal J_\star}{\partial T_i} =              \frac{\mathcal I^\star_i}{T_i} + \frac{j}{\kappa} \frac{\partial \mathcal I^\star_i}{\partial \mathcal G_\star}\frac{\partial \mathcal G_\star}{\partial t},
	\end{align}
\end{subequations}
where integer $\kappa$ controls the relative resolution of quadrature.  $\rbrac{\bar{\omega}_0,\bar{\omega}_1,\dots,\bar{\omega}_{\kappa_i-1},\bar{\omega}_{\kappa_i}}=\rbrac{1/2,1,\cdots,1,1/2}$ are the quadrature coefficients following the trapezoidal rule \cite{cite:trapezoidal_rule}. $i = (1, 2, \cdots, N)$ denotes the $i$-th piece and $j = (1, 2, \cdots, \kappa)$.
\paragraph{Actuator Constraints}
Inspired by \cite{cite:feasible_constraint}, velocity $p_{i}^{(1)}(t)$, body rate $\omega$ and thrust $\tau$ of Eq. \ref{eq:actuator_constrain} can be constrained by constructing such a penalty function as follow, and the penalty gradients are caculated by
\begin{subequations}
	\label{eq:actuator_penalty}
	\begin{align}
		~\mathcal G_D(t) = \mathcal L_\mu [\norm{D(t)}^2-D_{max}^2], 
		D= p_{i}^{(1)}(t), \omega, \tau, \\
		\mathcal L_\mu[x] = 
		\begin{cases}
			0,                    & x \leq 0,       \\
			(\mu - x/2)(x/\mu)^3, & 0 < x \leq \mu, \\
			x - \mu/2,            & x > \mu.
		\end{cases} 
	\end{align}
	\vspace{-0.2cm}
	\begin{align}
		~\frac{\partial \mathcal G_D}{\partial c_i} =  2\beta _i^{(n)} (t) p_i^{(n)}(t)^T \frac{\mathcal L_\mu}{\partial x}, \\
		~\frac{\partial \mathcal G_D}{\partial T_i} = 2\beta _i^{(n)}(t) c_i p^{(n)}(t) \frac{\mathcal L_\mu}{\partial x}.
	\end{align}
\end{subequations}

\paragraph{Safety Constraints}
The safety constraint Eq. \ref{eq:ground_constrain} avoiding the collision with ground can be transformed to such a penalty function \cite{cite:perching} as follow
\begin{equation}
	\begin{aligned}
		\label{eq:ground_penalty}
		\mathcal G_g(t) = \mathcal L_\mu [z_{min}^2 - \norm{\mathbf e_3^T p(t)}^2].
	\end{aligned}
\end{equation}
And the gradients are calculated just like Eq. \ref{eq:actuator_penalty}.

Summarizing the above strategies, the constraints of the Eq. \ref{eq:problem_formulation} are unified to the same unconstrained cost function Eq. \ref{eq:nonlinear_const_function}. The requiements can be trade off by adjusting the weight vector $\lambda$. We set the initial values according to $p_0(t)$ in Section \ref{section:primitive} by calculating each intermediate waypoint position using the uniform time from decision making module. The problem is then efficiently solved by the L-BFGS \cite{cite:LBFGS} whose solver is open source\footnote{https://github.com/ZJU-FAST-Lab/LBFGS-Lite}.

\section{Experiments}
\begin{figure*}[htbp]
	\subfloat[The process of the quadrotor catching triple targets continuously]{
		\includegraphics[width=2.0\columnwidth]{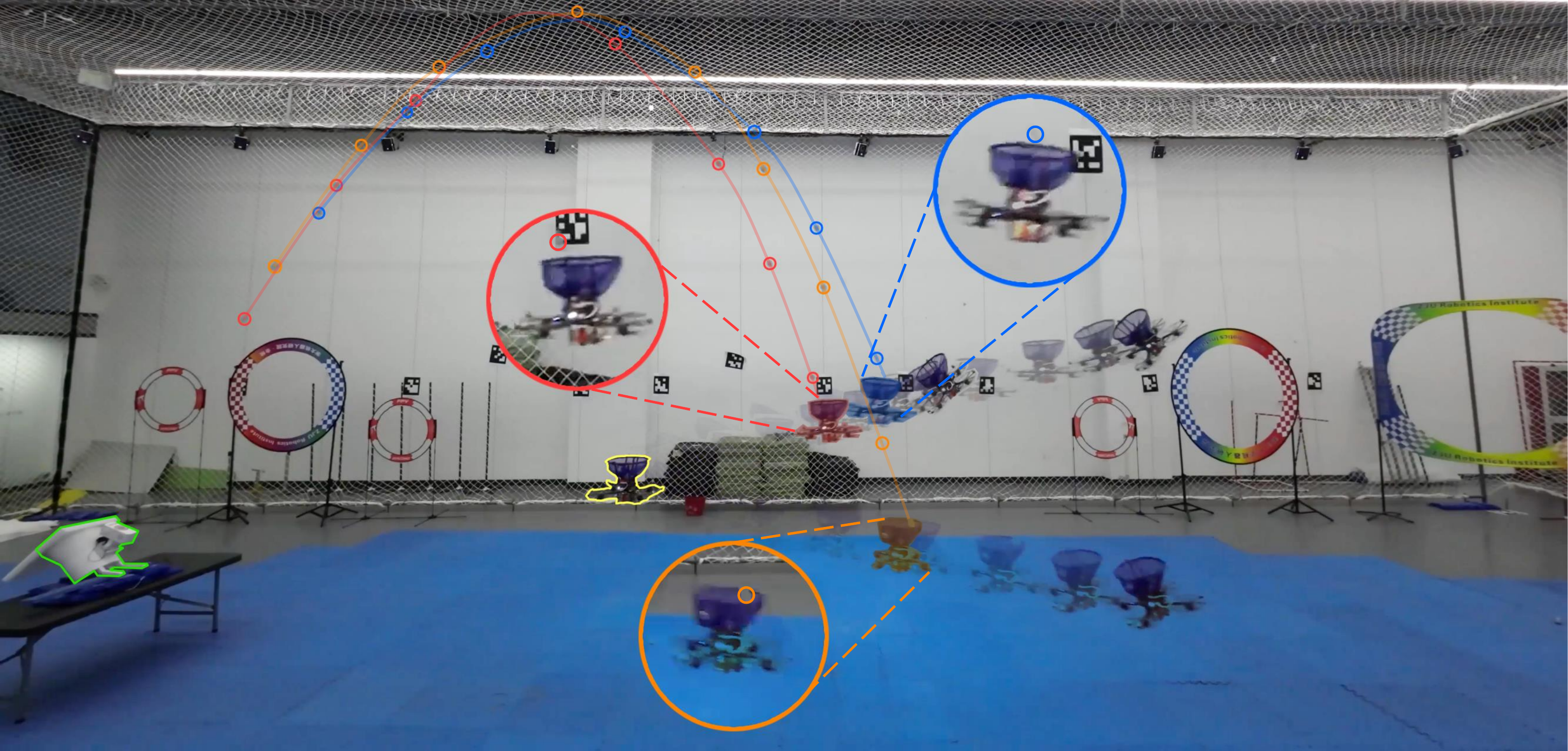}
		\label{fig:triple}
	}
	\vspace{-0.3cm}	
	\subfloat[Position map of ball catching process]{
		\includegraphics[width=0.9\columnwidth]{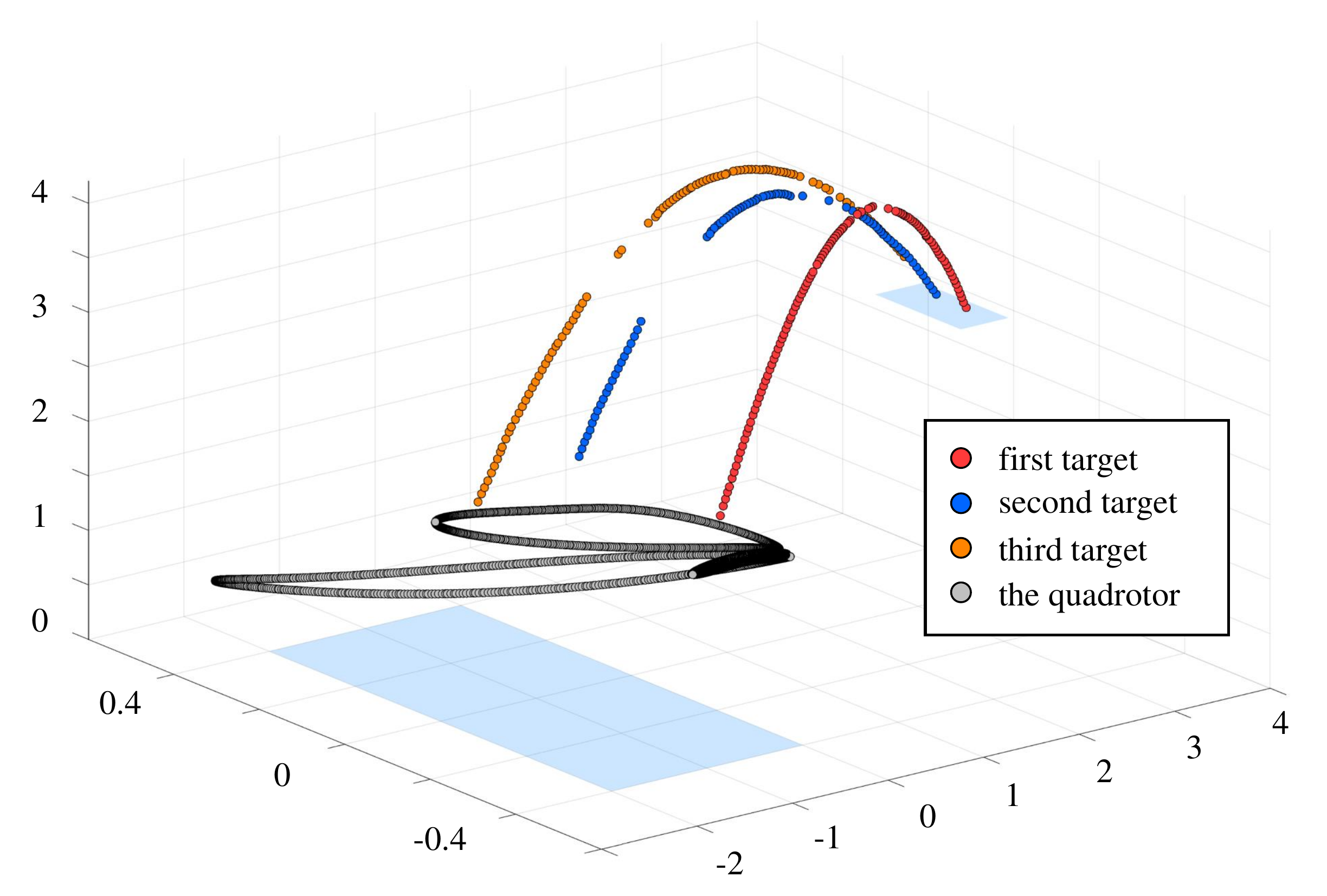}
		\label{fig:catching}
	}
	\subfloat[The quadrotor]{
		\includegraphics[width=0.52\columnwidth]{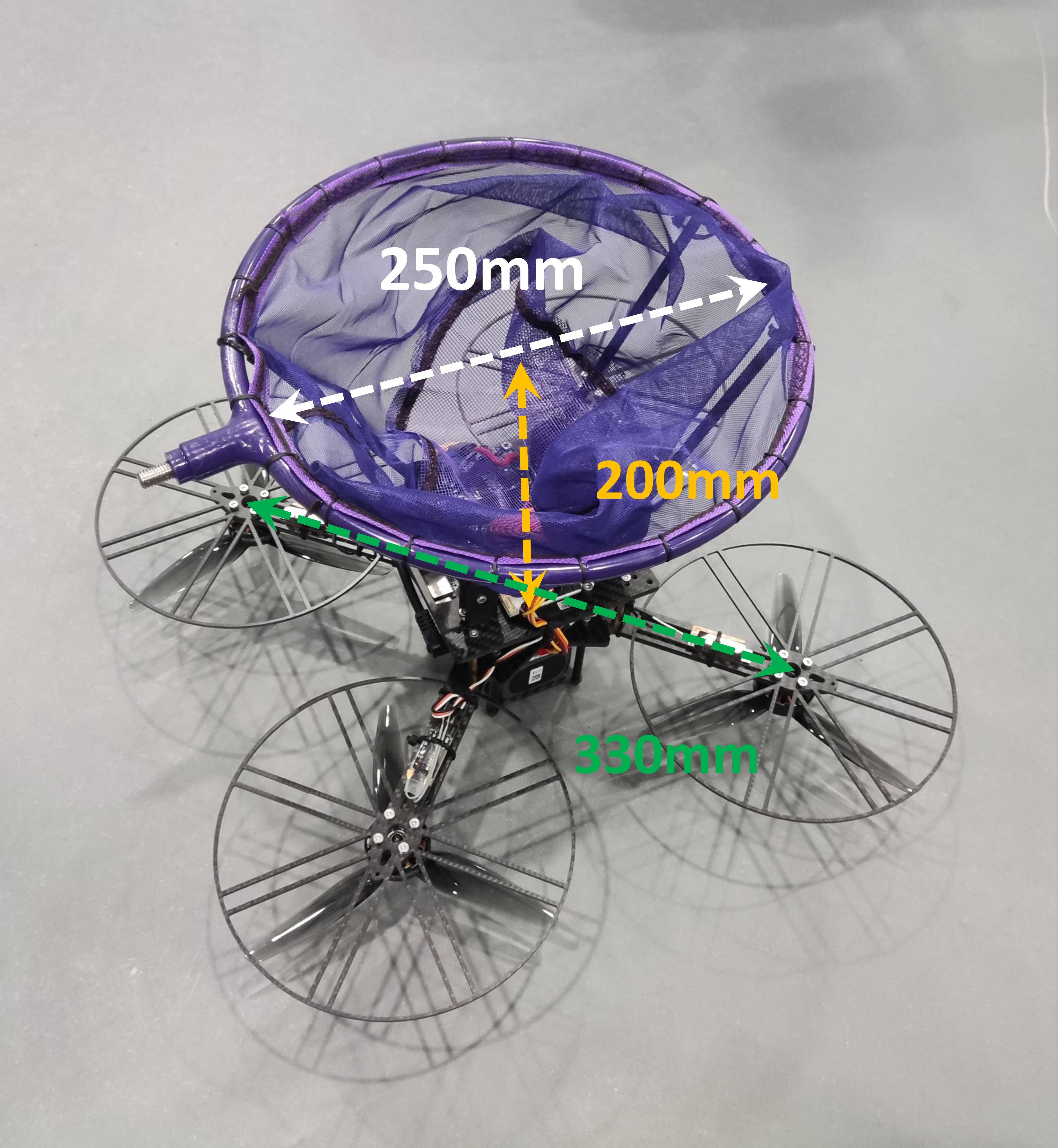}
		\label{fig:quadrotor}
	}
	\subfloat[The pitching machine]{
		\includegraphics[width=0.53\columnwidth]{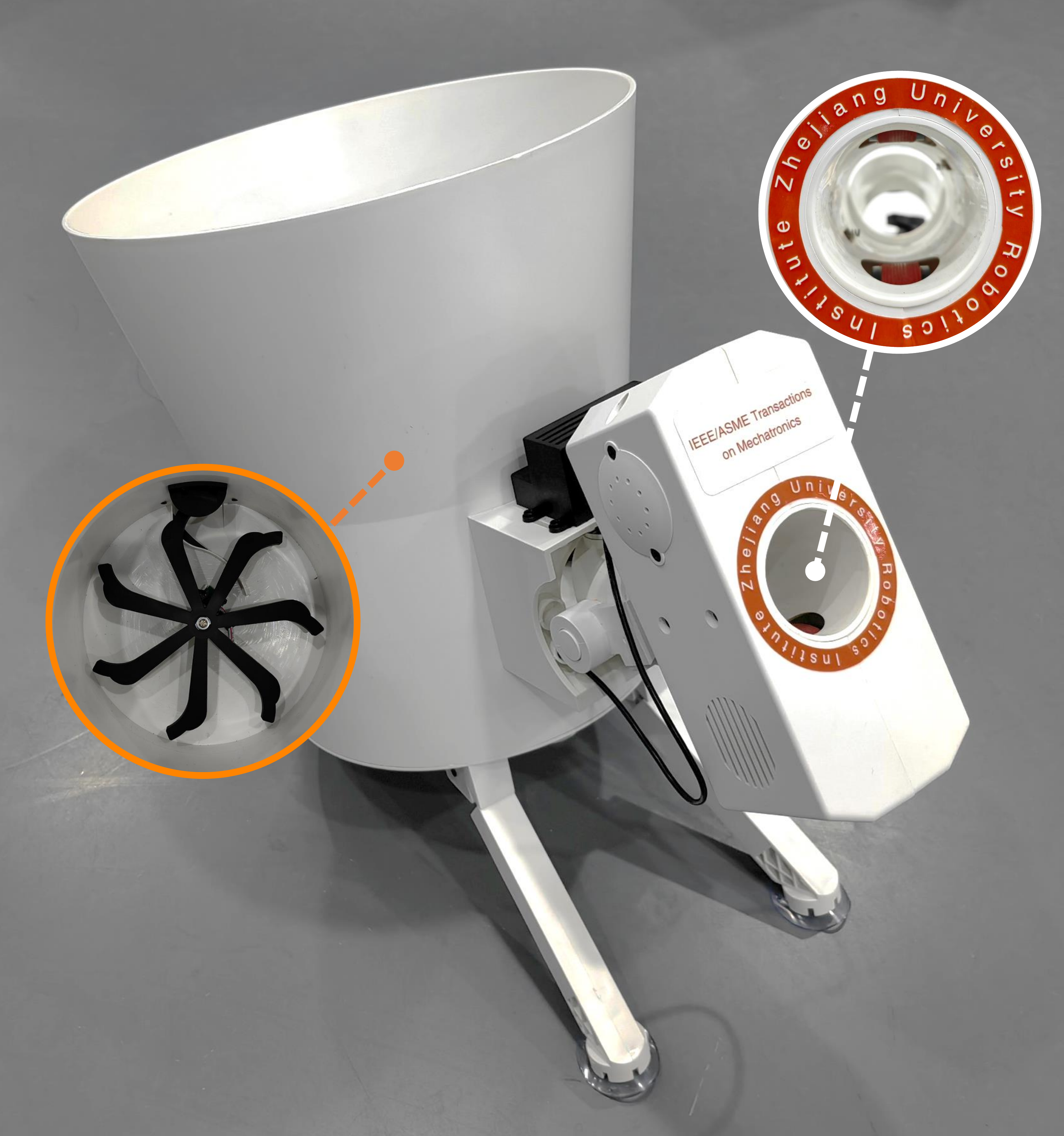}
		\label{fig:throwing}
	}	
	\vspace{0cm}	
	\subfloat[Velocity]{
		\includegraphics[width=0.49\columnwidth]{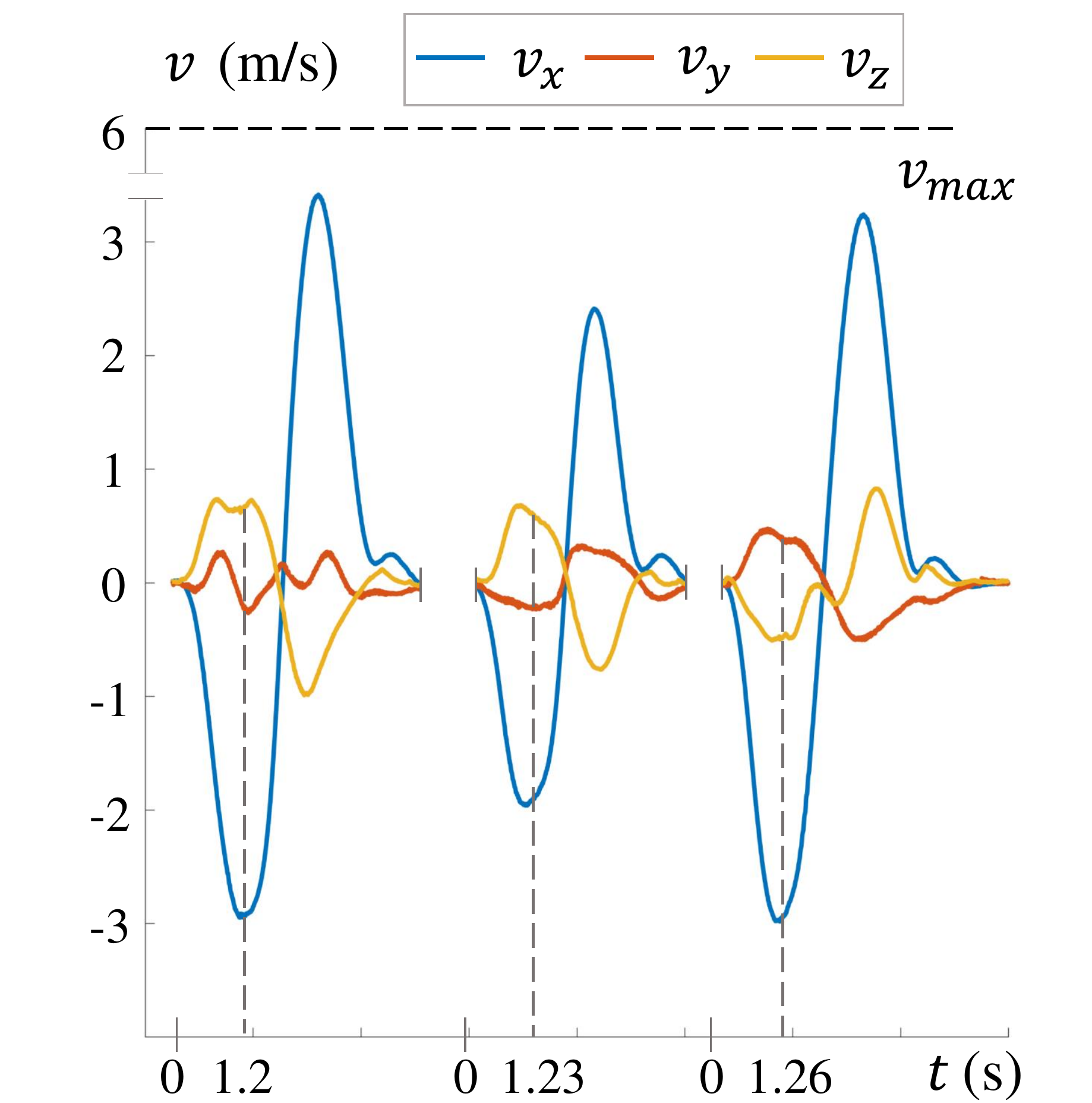}
		\label{fig:actuator_v}
	}
	\subfloat[Body Rate]{
		\includegraphics[width=0.49\columnwidth]{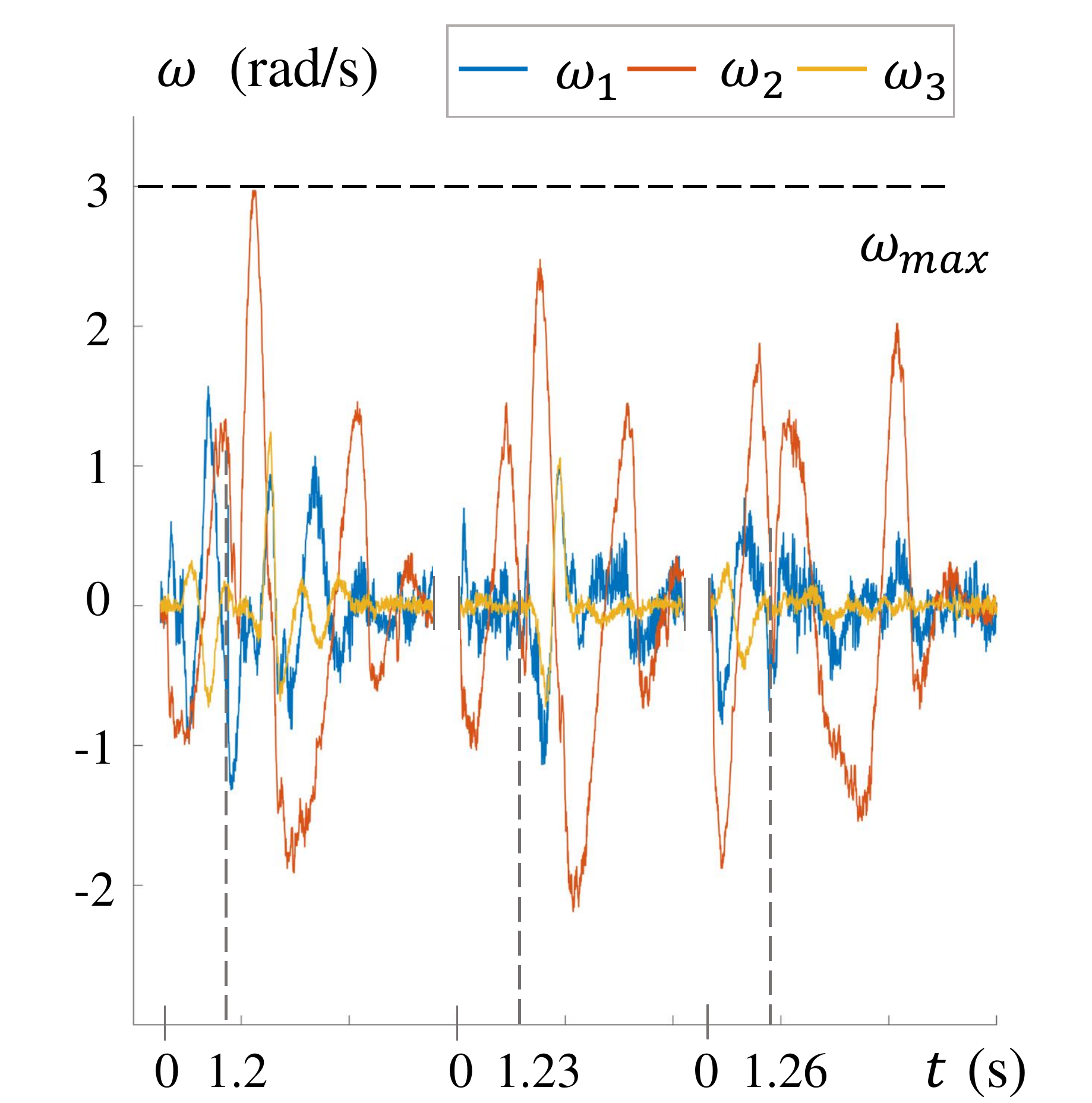}
		\label{fig:actuator_w}
	}
	\subfloat[Thrust]{
		\includegraphics[width=0.49\columnwidth]{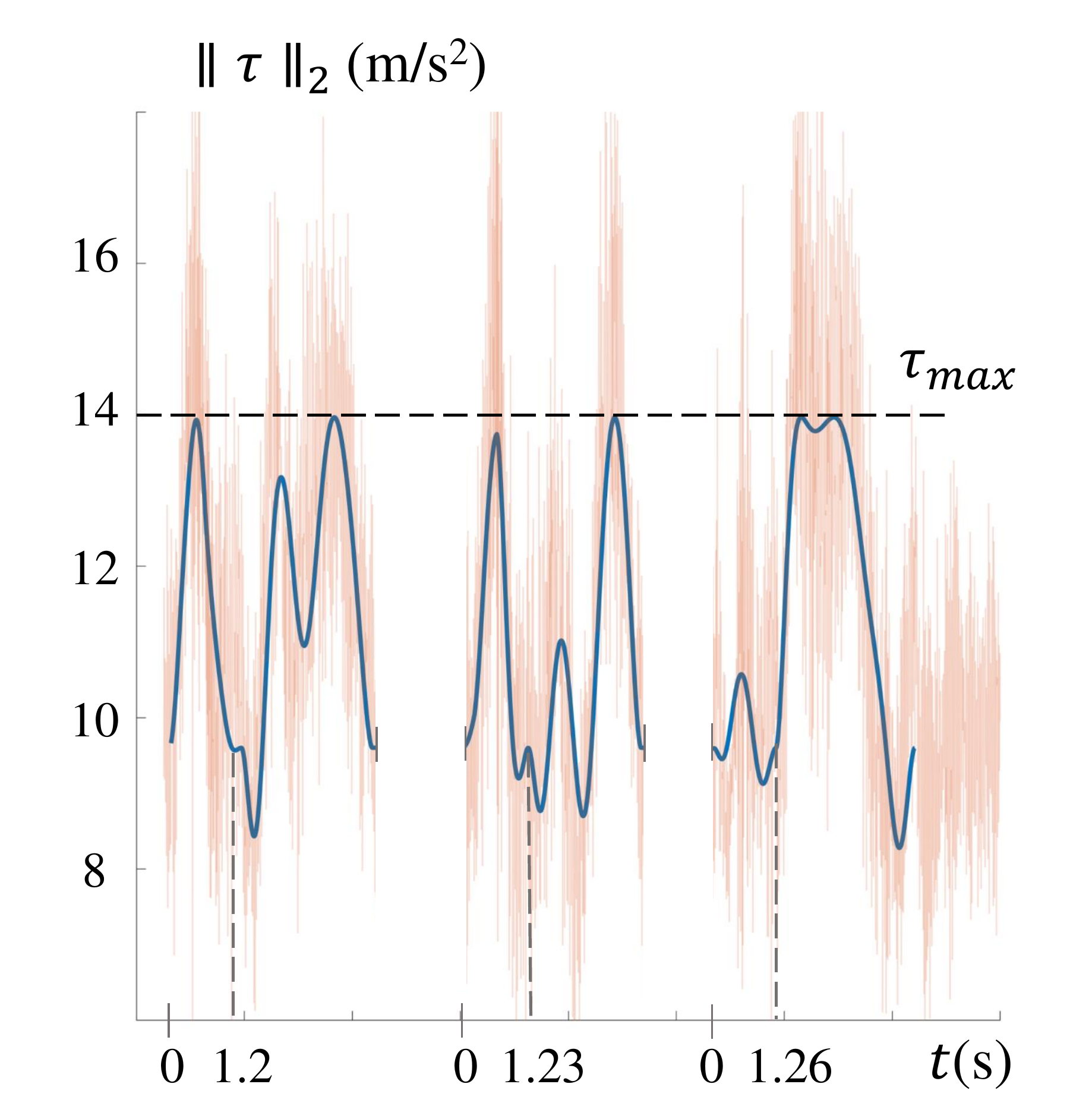}
		\label{fig:actuator_f}
	}
	\subfloat[Jerk]{
		\includegraphics[width=0.5\columnwidth]{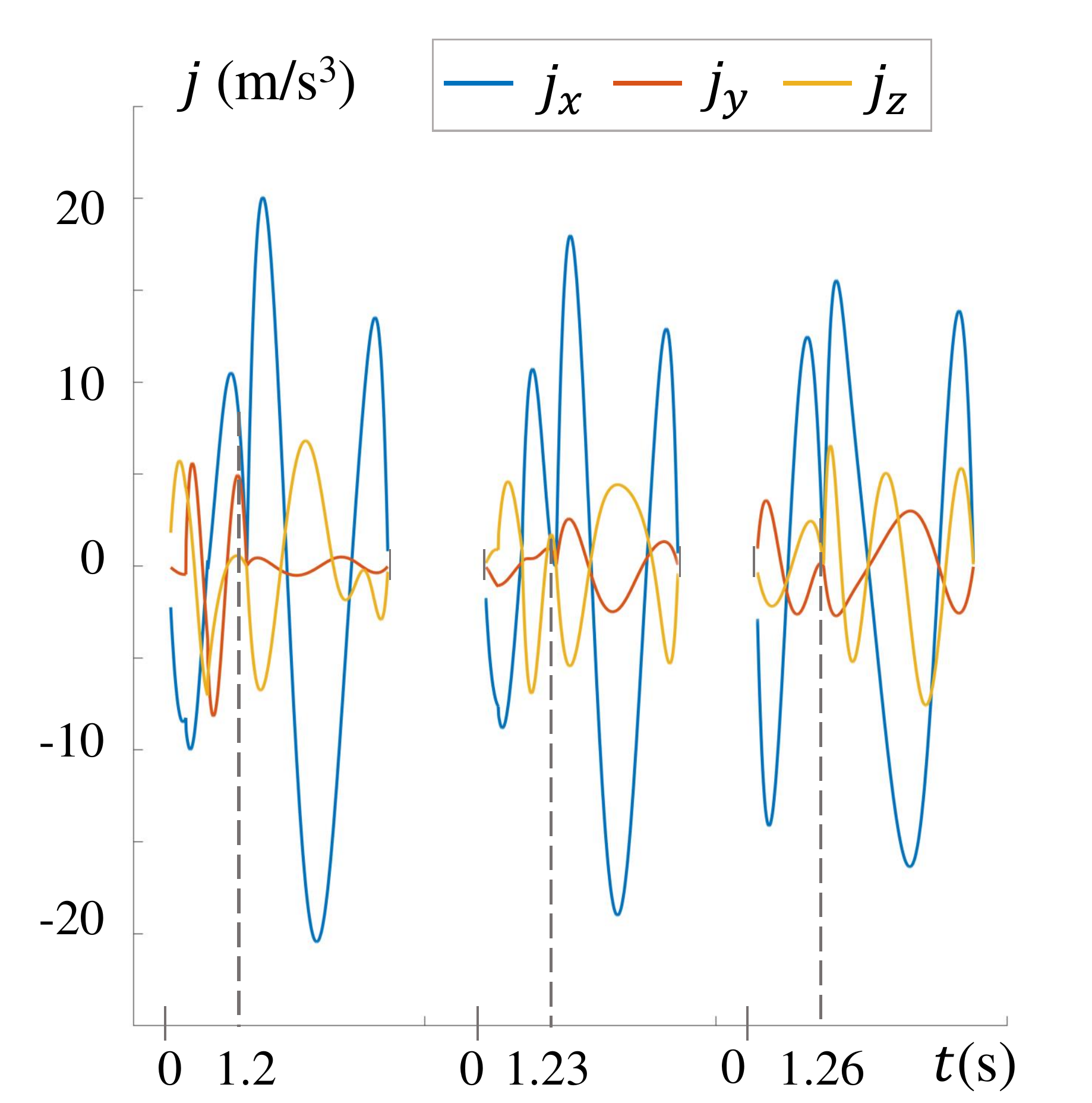}
		\label{fig:actuator_j}
	}
	\caption{
		Experimental platform and results. (a) shows the process of catching triple targets. The time from each target thrown to hitting the ground (if not catched) is less than 1.9$s$. The time left for the quadrotor response is less than 1.6$s$. The catching time is less than 1.4$s$. The motion trajectories of three targets are represented by red, blue and orange respectively. Local images are magnified to see the moment of catching the targets clearly. The trajectories of the quadrotor are represented by ghost images. The green box is the pitching machine. The yellow box is the starting position of quadrotor planning. (b) shows the positions of the quadrotor and the targets moving measured by Vicon cameras. The translucent rectangle is the erea where the ball is measured for the first time and the landing position. (c) shows the quadrotor designed by ourselves. The diameter of catching net is 250mm, and the diagonal propeller distance of the quadrotor is 330mm. The distance between the hoop plane and the propeller plane is 200mm. (d)shows the pitching machine. The ball is pushed to the tee by the wave wheel, and the ball is squeezed and launched by two rubber wheels driven by brushless motors. (e) shows the quadrotor velocity from an EKF. (f) shows the quadrotor angular rate measured by the IMU. Due to the vibration of the rack, the curve has a small jitter, which is a normal phenomenon. (g) shows the thrust. The orange translucent lines are the measured value with noise, and blue is the expected command. (h) shows the desired jerk.		
	}	
	\label{fig:experiment}
	\vspace{-0.5cm}
\end{figure*}

We design real scenes and simulated scenes to verify the robustness and extensibility of Catch Planner, and compare with the benchmark. 

\subsection{Real-world Experiments}
\subsubsection{Experiments Platform}
We design a quadrotor with a net attached above the center of mass, shown as Fig. \ref{fig:quadrotor}. An NUC is used as the onboard computer and all programs run on Intel I5-1135G7 CPU at 2.4GHz. The state-of-the-art Motion Primitive Planner (MPP)\cite{cite:bvp} is as the benchmark, which runs on the same computer. A $18m * 9m * 5m$ motion capture gym with 22 Vicon cameras is used as the experimental site, shown in Fig. \ref{fig:triple}. A pitching machine with two driven rubber wheels is used for throwing balls, shown as Fig. \ref{fig:throwing}. The state estimation of the quadrotor is given by an EKF of the pose from Vicon cameras and the IMU data from a PX4 Autopilot. We adopt the $SE(3)$ casecade PID controller using Hopf fibration \cite{cite:Holf} to avoid singularities and align the attitude calculation of planning and control. Control command is calculated from the trajectory by using differential flatness output model \cite{cite:minsnap}. 

The target throwing position is set at $(4, 0, 0.8)m $ under world coordinate. Because the throwing is random, the ball's landing position is within $2m \times 0.8m$ at the floor, shown in Fig. \ref{fig:catching}. Due to the height limitation, the balls only fly below 4.2$m$, which means that the flight time of the ball does not exceed 1.9$s$. In fact, in our experiment, the sensing data is considered stable and used only after the balls are thrown higher than 2$m$. After the first target motion model updated, historical data will be temporarily stored for optimizing the drag coefficient, which is output after the ball exceeds the height of 2.8$m$. In this case, the whole flight time of each target does not exceed 1.6$s$. 

\subsubsection{Experimential Results}
\begin{figure}[t]
	\begin{center}
		\subfloat[Velocity]{
			\includegraphics[width=0.3\columnwidth]{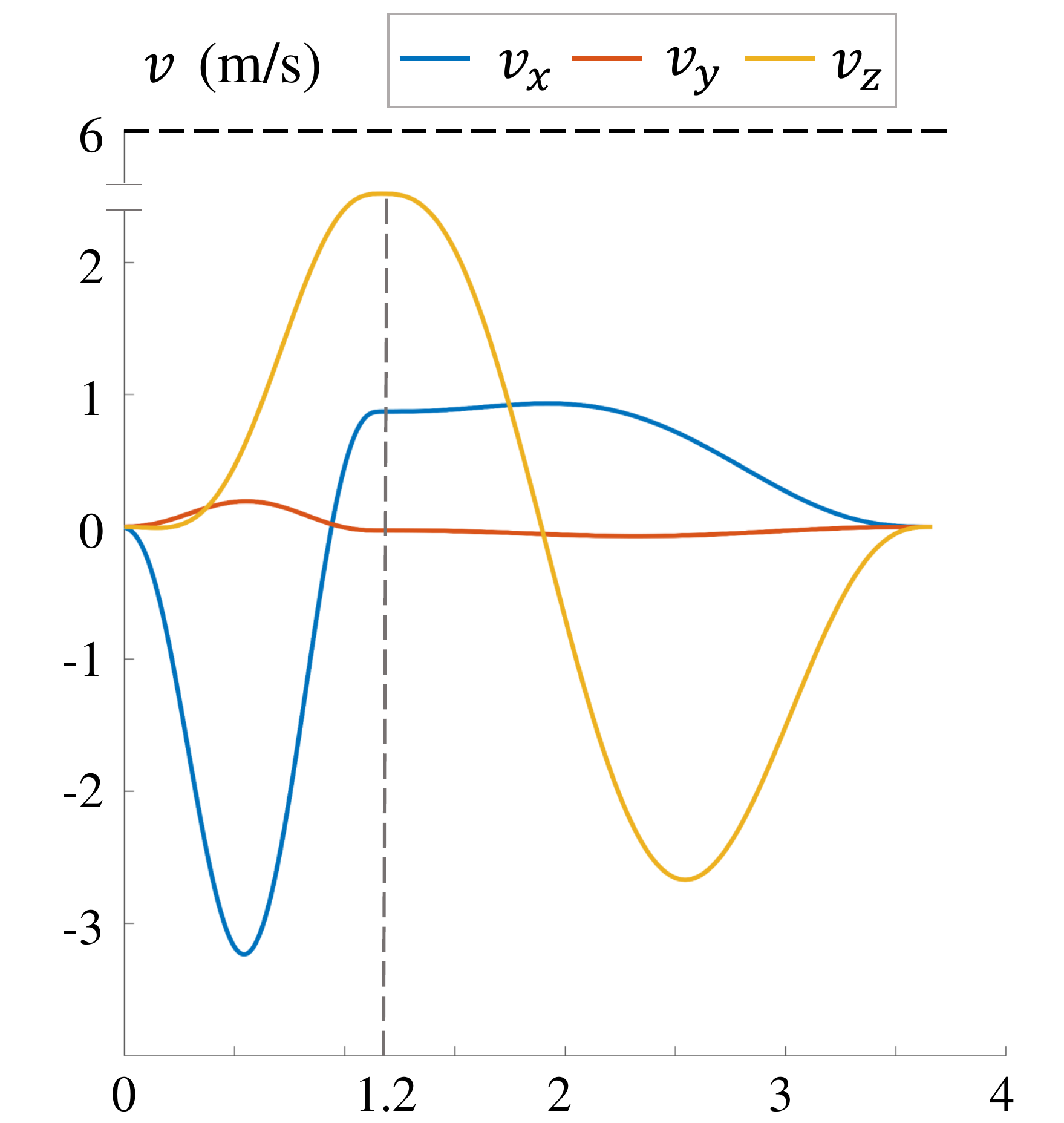}
		}
		\subfloat[Body Rate]{
			\includegraphics[width=0.3\columnwidth]{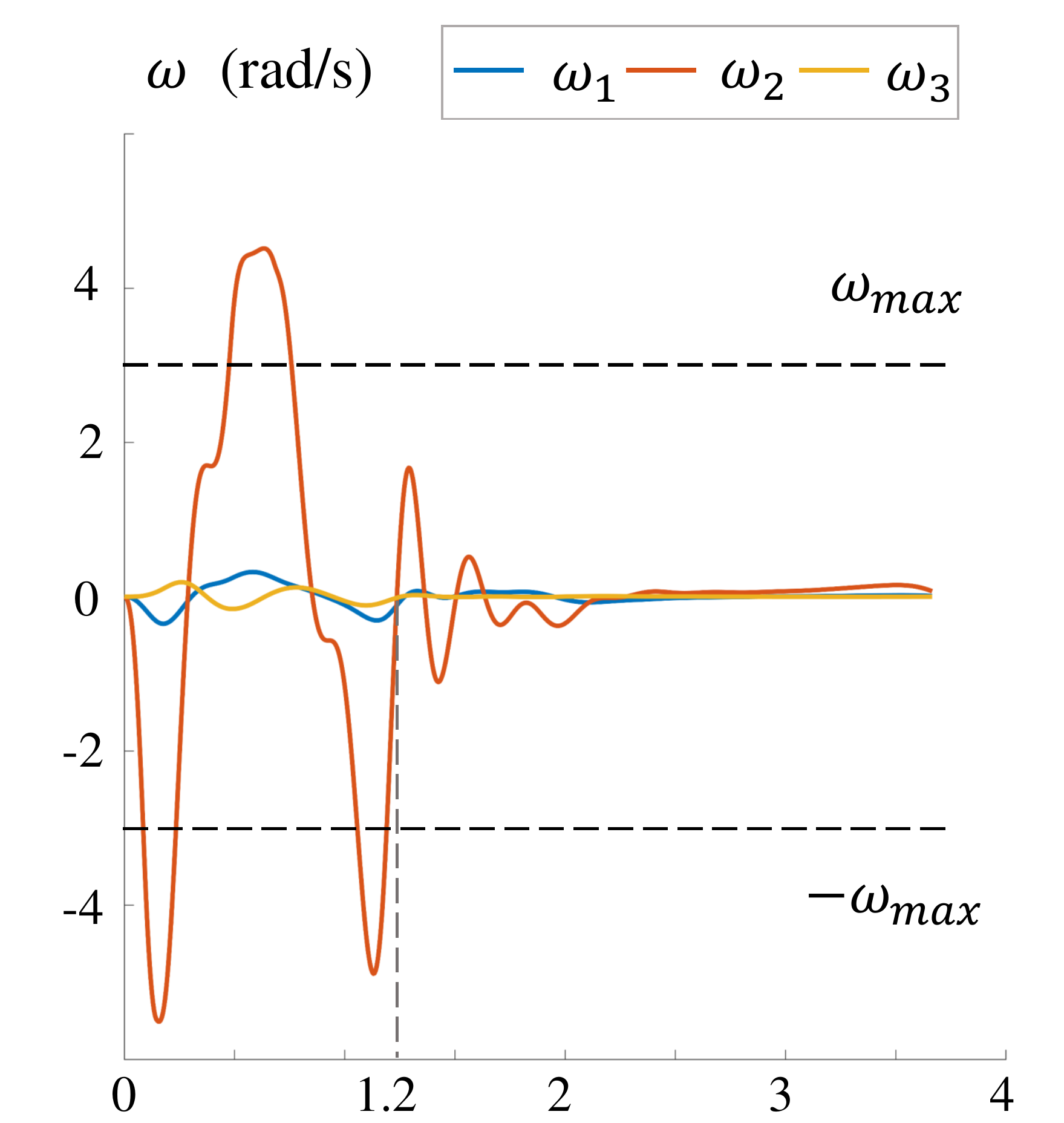}
		}
		\subfloat[Thrust]{
			\includegraphics[width=0.3\columnwidth]{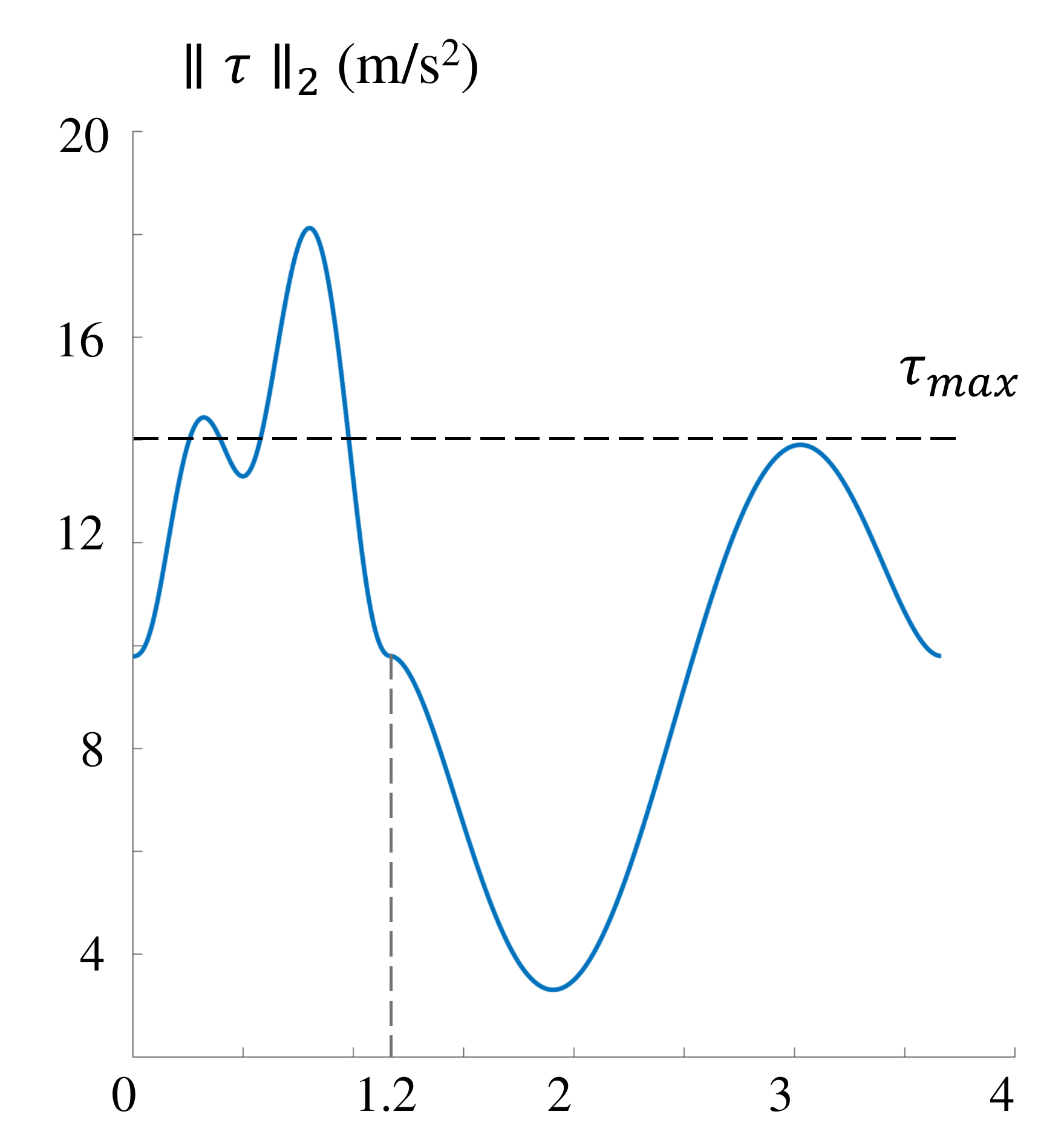}
		}
	\end{center}
	\caption{
		\label{fig:benchmark}
		The planning dynamics of the benchmark. 
	}
	\vspace{-0.5cm}
\end{figure}

We check the dynamic feasibility, safety and catching constraints of the desired trajectory to ensure that the calculated trajectory is executable. If the trajectory passes the check, the planning-with-decision is considered successful. If the ball falls into the net, it will be regarded as a successful catching. In addition, we propose the optimal time ratio (OTR) to measure the gap between decision results and optimal results. The MPP has no OTR because it has no decision making ability. OTR is defined as follows:
\begin{align}
	OTR = \frac{1}{n}\sum \frac{|T_o-T_p|}{T_o},
\end{align}
which measures the proximity of $n$ outputs of the decision making module $T_p \in \mathcal T_p$ and optimized time $T_o \in \mathcal T_o $. Obviously, the closer the decision result is to the optimal result, the smaller OTR is. It also provides a pattern for measuring the ability of decision making under similar planning-with-decision frameworks. 

We carry out 50 experiments respectively for Catch Planner and MPP whose terminal catching height is set to 2$m$. Just like \cite{cite:bvp}, the terminal velocity and acceleration are sampled, and the trajectory with the lowest cost is finally adopted. Table \ref{tab:real-world_result} summarizes the experimental results. It can be seen that the success rate (SR) of Catch Planner is far greater than that of MPP. Due to the gap between ideal and reality, the success rate of catching will never be higher than the success rate of planning.
\begin{table}[ht]
	\renewcommand\arraystretch{1.2}
	\centering
	\caption{Success Rate and Optimal Time Ratio}
	\label{tab:real-world_result}
	\begin{tabular}{ccccc}
		\toprule
		\multicolumn{2}{c}{}                          &  \makecell{Planning 
			SR} & \makecell{Catching 
			SR} & OTR    \\
		\midrule
		\multicolumn{2}{c}{Catch Planner}    & 96\%    & 64\%     &  0.092        \\
		\multicolumn{2}{c}{MPP\cite{cite:bvp}}      & 14\%    & 8\%     & /        \\
		\bottomrule
	\end{tabular}
\end{table}

Three consecutive successful catch cases are selected to analyze the dynamics. We only visualize the real data of the quadrotor during moving, not including hovering. Fig. \ref{fig:experiment}e-g show that the planned trajectory can effectively constrain the velocity, angular velocity and thrust. Fig. \ref{fig:actuator_j} shows that the planned trajectory is smooth enough, even jerk. The logged Rosbag of the first catching in Fig. \ref{fig:experiment} is used to simulate the same throwing. The catching time is set to be the same, which means the same state of caught targets. The desired trajectory of MPP is shown in the Fig. \ref{fig:benchmark}. It can be seen that ignoring the dynamic feasibility and safety of the benchmark leads to some actuator's command exceeding the limit, which often appears in the 50 experiments.

Our method can also catch multiple flying balls, which benefits from the real-time decision making ability. Although it is within the 4.2$m$ height limit, the quadrotor still shows the ability to catch two targets flying together in the air, as shown in Fig. \ref{fig:introduction}.

\begin{table}[ht]
	\renewcommand\arraystretch{1.2}
	\centering
	\caption{Characteristics Comparison}
	\label{tab:characteristics}
	\begin{tabular}{ccccc}
		\toprule
		\multicolumn{2}{c}{}                          & Catch Planner     & MPP \cite{cite:bvp}         \\
		\midrule
		\multicolumn{2}{c}{Decision-making Ability}      &  \textbf{Autonomous}       & Manual  \\		\multicolumn{2}{c}{Time \& Terminal State}    & \textbf{Optimize}        &  Fix        \\
		\multicolumn{2}{c}{Dynamic Feasibility}      &  \textbf{Optimize}           & Check          \\
		\multicolumn{2}{c}{Collision Safety}      &  \textbf{Optimize}           & Check         \\
		\multicolumn{2}{c}{Computing  Consumption}      & 9 $ms$       &  \textbf{0.149 $ms$}         \\
		\bottomrule
	\end{tabular}
\end{table}

The characteristics of Catch Planner and MPP are summarized as the Table \ref{tab:characteristics}. MPP does not have high-level decision making ability, but relies on manual assignment. The state of the target is highly coupled with the time. This is not considered by MPP \cite{cite:bvp}. The dynamic feasibility and safety are also ignored during planning and only checked the end state. In addition, For computing consumption, although the calculted time of a single trajectory in \cite{cite:bvp} is far less than our method, 9$ms$ is enough for the 100$hz$ frequency of state estimation and control, while only 2$ms$ is required for re-planning.

\subsection{Simulation Experiments }
\subsubsection{Scene Simulation}
We design 4 scenarios to evaluate our planning-with-decision method according to different target trajectory types. Compared with real experiments, simulation scenarios effectively eliminate the target prediction error, sensing error, control error and communication delay in the real environment, so that we can focus on verifying the effectiveness of planning-with-decision methods. Thanks to the economy of simulation, we test 1000 experiments for each scenario  in short time. The different trajectories types of the targets are as follow:

\paragraph{Parabola} This simulation to throw 2 targets is highly consistent with the real experimental scene, except that the target is free from height limitation and wind resistance. The change of position from initial position $p_0$ can be expressed by acceleration $a$ and velocity $v_{pa}$. The motion trajectory is as $p_{pa}(t) = 1/2at^2+v_{pa} t+p_0,\  t<2$.
\paragraph{Harmonic} 2 targets move at the speed $v_{ha}$ in the shape of harmonic, which are expressed as $p_{ha}(t) = sin(v_{ha}t) +p_0,\ t<7$.
\paragraph{Triangle} 3 diagonally moving objects move and form an equilateral triangle. For the $i$-th and $j$-th targets, the position can be calculated as $p_{tr}(t) = v_{tr}t+p_0, \ \left< v^i, v^j \right>=2/3\pi \ and \ v^i \neq v^j for \ \forall v^i, v^j,\  t<6$.
\paragraph{Hexagon} The position of 6 targets flying in parallel at a constant velocity $v_{he}$ can be expressed as $p_{he}(t) = v_{he}t+p_0,\  t<10$.

It is worth noting that the above trajectories are all on $e_3$. The target keeps constant velocity on $e_1$ and $e_2$. 
\begin{figure}[t]
	\begin{center}
		\includegraphics[width=0.8\columnwidth]{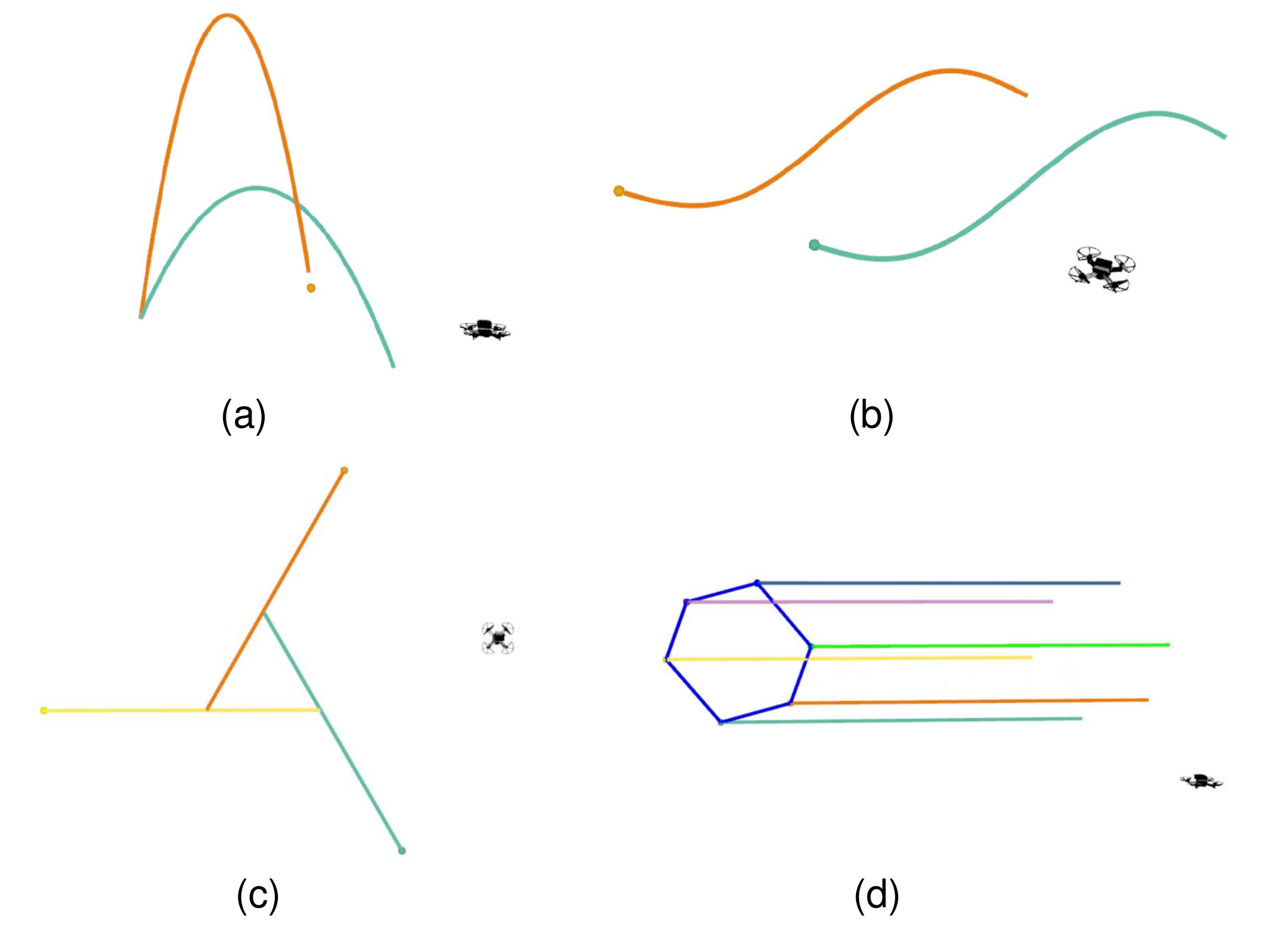}
	\end{center}
	\vspace{-0.2cm}
	\caption{
		\label{fig:simulation}
		The simulation scenarios. Targets flight paths are visualized as curves with multiple colors. (a) shows Parabola. (b) shows Harmonic. (c) shows Triangle. (d) shows Hexagon.
	}
	\vspace{-0.8cm}
\end{figure}
\subsubsection{Experimential Results}
In our simulation experiments, there is no target trajectory prediction and quadrotor control error. The success rate and optimal time ratio in Table \ref{tab:simulation_result} can reflect the ability of our method through thousands of experiments. 

The results show that the catching difficulty increases as the targets amount increases and trajectory becomes more complex. We also compare different state-of-the-art RL methods for continuous tasks including PPO \cite{cite:ppo}, SAC \cite{cite:sac} and TQC \cite{cite:tqc}. PPO  outperforms other methods in complex scenarios while performs similarly in simple scenarios. 

\begin{table}[ht]
	\renewcommand\arraystretch{1.2}
	\setlength{\tabcolsep}{0.15cm}
	\centering
	\caption{Success Rate and Optimal Time Ratio}
	\label{tab:simulation_result}
	\begin{tabular}{ccccccc}
		\toprule
		\multicolumn{2}{c}{Scene}      & \makecell{Parabola \\ (1 target)  }   & \makecell{Parabola \\ (2 targets)  } & 		\makecell{Harmonic \\ (2 targets)}  & \makecell{Triangle \\ (3 targets)}   & \makecell{Hexagon \\ (6 targets)}  \\
		\midrule
		\multicolumn{2}{c}{\makecell{PPO SR} }    & 99.3\%  & 98.2\%  & 84.7\%  & 99.7\%  & 76.3\%      \\
		\multicolumn{2}{c}{PPO OTR}      		  &0.052  & 0.119  & 0.234  & 0.126  & 0.458      \\
		\multicolumn{2}{c}{\makecell{
				SAC SR} }    & 99.8\%  & 99.6\%  & 75.7\%  & 99.8\%  & 65.7\%      \\
		\multicolumn{2}{c}{SAC OTR}      		  &0.059  & 0.194  & 0.316  & 0.137  & 0.484      \\
		\multicolumn{2}{c}{\makecell{TQC SR} }    & 99.9\%  & 99.7\%  & 68.5\%  & 99.9\%  & 69.6\%      \\
		\multicolumn{2}{c}{TQC OTR}      		  &0.041  & 0.215  & 0.373  & 0.122  & 0.491      \\
		\bottomrule
	\end{tabular}
\end{table}

The experimental results demonstrate that our method can be extended to catch targets of various trajectories. In fact, the only requirement for using our method is that the motion of targets can be differentially analytically expressed. This condition can be easily achieved by establishing curve fitting or simplifying dynamic models. In addition, experience tells us that the success rate usually exceeds 80\% when the OTR is below 0.3. Although it has not been proved by theory, it can be used as a reference for researchers. 

\section{Conclusion}
In this paper, we propose a novel planning-with-decision framework Catch Planner to catch high-speed moving targets when facing \textbf{F}ormulistic, \textbf{L}ightweight, \textbf{A}daptive and \textbf{F}lexible requirements (FLAF). It integrates decision making, motion primitive generation, trajectory optimization, and target trajectory prediction modules. 

Under Catch Planner, the advantages of learning based and optimization based methods complement each other by the coupling of motion planning and decision making, meeting the \textbf{F}ormulistic need. We propose a DRL based policy search method for decision making and a self-supervised neural network training method. Then, we propose a terminal-flexible optimal trajectory optimization method for optimal catching. Facing different flying targets, the coupling time and terminal position are jointly optimized to \textbf{A}daptively catch, and a terminal constraint transformation method is proposed to make the catching \textbf{F}lexible. All the above are solved online with the calculation consumption does not exceed 10$ms$, which proves the \textbf{L}ightweight of the framework.

We conduct simulations and real experiments to verify the effectiveness and robustness of the proposed method. The real world planning success rate is 96\%. The universality of the algorithm is verified by setting different target trajectories. Furthermore, this method has the potential to solve other highly dynamic problems. 

\section{Acknowledgment}
The authors would like to thank Prof. Hao Li and Li Xu for their valuable suggestions.

\end{document}